\documentclass[letterpaper]{article} 
\usepackage[]{aaai24}  
\usepackage{times}  
\usepackage{helvet}  
\usepackage{courier}  
\usepackage[hyphens]{url}  
\usepackage{graphicx} 
\def\UrlFont{\rm}  
\usepackage{natbib}  
\usepackage{caption} 
\usepackage{algorithm}
\usepackage{algorithmic}

\usepackage{float}
\usepackage[utf8]{inputenc}
\usepackage{amsmath}

\usepackage{amssymb}
\usepackage[capitalise]{cleveref}
\usepackage{multicol}
\usepackage{multirow}
\usepackage{graphicx}
\usepackage{amsthm}

\DeclareMathOperator*{\E}{\mathbb{E}}
\DeclareMathOperator*{\MI}{\mathcal{I}}
\DeclareMathOperator*{\MU}{\mathcal{U}}

\newtheorem{lemma}{Lemma}
\newcommand{\blemma}{\begin{lemma}}
\newcommand{\elemma}{\end{lemma}}

\newtheorem{thm}{Theorem}
\newcommand{\bthm}{\begin{thm}}
\newcommand{\ethm}{\end{thm}}

\newtheorem{defin}{Definition}
\newcommand{\bdefin}{\begin{defin}}
\newcommand{\edefin}{\end{defin}}

\newtheorem{observation}{Observation}
\newcommand{\bobserv}{\begin{observation}}
\newcommand{\eobserv}{\end{observation}}

\newtheorem{coroll}{Corollary}
\newcommand{\bcoroll}{\begin{coroll}}
\newcommand{\ecoroll}{\end{coroll}}

\newcommand{\bproof}{{\bf Proof Sketch: }}
\newcommand{\eproof}{\mbox{$\Box$}}

\usepackage{newfloat}
\usepackage{listings}
\DeclareCaptionStyle{ruled}{labelfont=normalfont,labelsep=colon,strut=off} 
\floatstyle{ruled}
\newfloat{listing}{tb}{lst}{}
\floatname{listing}{Listing}
\title{Revisiting Recommendation Loss Functions through Contrastive Learning (Technical Report)}

\author{
    Dong Li \textsuperscript{\rm 1},
    Ruoming Jin \textsuperscript{\rm 1},
    Bin Ren \textsuperscript{\rm 2}
}
\affiliations{
    \textsuperscript{\rm 1}Kent State University\\
    \textsuperscript{\rm 2} William \& Mary\\
dli12@kent.edu, rjin1@kent.edu, bren@cs.wm.edu

}

\begin{document}

\maketitle

\begin{abstract}

Inspired by the success of contrastive learning, we systematically examine recommendation losses, including listwise (softmax), pairwise (BPR), and pointwise (MSE and CCL) losses. In this endeavor, we introduce InfoNCE+, an optimized generalization of InfoNCE with balance coefficients, and highlight its performance advantages, particularly when aligned with our new decoupled contrastive loss, MINE+. We also leverage debiased InfoNCE to debias  pointwise recommendation loss (CCL) as Debiased CCL. Interestingly, our analysis reveals that linear models like iALS and EASE are inherently debiased. Empirical results demonstrates the effectiveness of MINE+ and Debiased-CCL. 
\end{abstract}

\maketitle

\section{Introduction}
Recommendation and personalization have become fundamental topics and applications in our daily lives \cite{charubook,zhang2019deep}. Recent research has demonstrated the difficulty of properly evaluating recommendation models~\cite{RecSys19Evaluation}, even for the (simple) baselines ~\cite{difficulty@rendle}.  A few latest studies have shown the surprising effectiveness of simple linear models and losses, such as iALS~\cite{ials_revisiting,ials++} and SimpleX (CCL)~\cite{simplex}. Concurrently, recommendation research is placing greater emphasis on the need for a deeper (theoretical) understanding of recommendation models and loss functions ~\cite{jin@linear}.

Drawing inspiration from recent studies, this paper delves deep into an analysis and comparison of loss functions, pivotal in developing recommendation models~\cite{simplex}.  In the past few years,  {\em contrastive learning}~\cite{SimCLR,infonce,debiased} has shown great success in various learning tasks, such as computer vision and NLP, among others in both supervised and unsupervised settings. It aims to learn a low dimensional data representation so that similar data pairs stay close to each other while dissimilar ones are far apart. Notably, contrastive loss functions (such as InfoNCE) ~\cite{SimCLR,infonce}  share striking resemblances with recommendation loss functions, like BPR (Bayesian Personalized Ranking) and Softmax~\cite{charubook}. The latter aims to align user embeddings more closely with the items they favor. The burgeoning (theoretical) exploration in contrastive learning presents an opportunity to deepen our understanding of recommendation losses and design new ones through the perspective of contrastive losses. Interestingly, recent studies~\cite{cl-rec,alignment-uniformity} have begun integrating contrastive learning approaches into individual recommendation models. Yet, a comprehensive comprehension of recommendation loss functions viewed through the prism of contrastive learning remains to be fully developed and explored.




Specifically, the following key research questions remain unanswered: How do listwise softmax loss and pairwise BPR~\cite{bpr} (Bayesian Pairwise Ranking) loss relate to these recent contrastive losses~\cite{debiased,infonce,MINE}?Can we design better recommendation loss functions by taking advantage of latest contrastive losses, such as InfoNCE~\cite{infonce}, decoupled ~\cite{decoupling@eccv} and mutual information-based losses\citep{MINE}? Can we {\em debias} $L1$ (mean-absolute-error, MAE) and $L2$ (mean-squared-error, MSE) loss function similar to debiasing contrastive loss, InfoNCE ~\cite{debiased}? More specific,  do the well-known linear models (based on $L2$ loss), such as iALS \citep{ials_revisiting,hu2008collaborative,ials++} and EASE \citep{ease}, need to be debiased with respect to contrastive learning losses? 


To answer all these questions, we made the following contributions in this paper: 
\begin{itemize}
\item We revisit commonly used contrastive learning loss - InfoNCE in the recommender system and propose its general format - InfoNCE+ by introducing balance coefficients on the denominator as well as the positive term inside. Our empirical results appeal that the discarding of the positive term in the denominator would lead to superior performance which aligns with the latest decoupled contrastive loss ~\cite{decoupling@eccv}. We further theoretically explore it through mutual information theory and simplify it as MINE+. Finally, through the lower bound analysis, we are able to relate the contrastive learning losses and BPR loss. 

\item By leveraging the latest debiased contrastive loss~\cite{debiased}, we propose debiased InfoNCE and generalize the debiased contrastive loss ~\cite{debiased} to the pointwise loss CCL~\cite{simplex}) in recommendation models, referred as Debiased CCL. We then examine the well-known linear models with pointwise $L2$ loss, including iALS and EASE; rather surprisingly, our theoretical analysis reveals that both are inherently debiased under the reasonable assumption with respect to the contrastive learning losses.    
\item We experimentally validate the debiased InfoNCE and point-wise losses indeed perform better than the biased (commonly used) alternatives. We also show the surprising performance of the newly introduced mutual information-based recommendation loss and its refinement (MINE+). 
 \end{itemize}

To the best of our knowledge, this is the first study to systematically investigate the recommendation loss functions through the lens of contrastive learning. It not only helps better understand and unify the recommendation losses but also introduces new loss functions, such as MINE+ and Debiased CCL, which are inspired by contrastive learning and introduced to the recommendation realm for the first time. Moreover, we establish the unexpected debias properties of iALS and EASE, unveiling significant characteristics previously unexplored.  These findings serve to construct a foundational (theoretical) scaffold for better understanding of recommendation losses.


\vspace*{-3.0ex}
\section{Background}
\label{sec:preliminary}
 In this section, we briefly review the commonly-used recommendation loss function - BPR (Bayesian Personalized Ranking) ~\cite{bpr}, softmax ~\cite{youtube,cl-rec}, and InfoNCE \cite{infonce}.

\noindent{\bf Notation:} 
We use $\mathcal{U}$ and $\mathcal{I}$ to denote the user and item sets, respectively. For an individual user $u\in \mathcal{U}$, we use $\mathcal{I}^+_u$ to denote the set of items that have been interacted (such as Clicked/Viewed or Purchased) by the user $u$, and $\mathcal{I}\backslash \MI^+_u$ to represent the remaining set of items that have not been interacted by user $u$. Similarly, $\MU^+_i$ consists of users interacting with item $i$, and $\MU\backslash \MU_i^+$ includes remaining users.
Often, we denoted $r_{ui}=1$ if item $u$ is known to interact with of item $i$ and $r_{ui}=0$ if such interaction is unknown.

Given this, most of the recommendation systems utilize either matrix factorization (shallow encoder) or deep neural architecture to produce a latent user vector $v_u$ and a latent item vector $v_i$, then use their inner product ($<v_u, v_i>$) or cosine ($<v_u/||v_u||, v_i/||v_i||>$) to produce a similarity measure $\hat{y}_{ui}$. The loss function is then used to produce a (differentiable) measure to quantify and encapsulate all such $\hat{y}_{ui}$. 

\noindent{\bf BPR Loss:}
The well-known pairwise BPR loss is the most widely used loss function in the recommendation, which can be written as:
\begin{equation}\label{eq:bpr}
\begin{split}
&{\mathcal{L}}_{bpr} = \E_u \E_{i\sim p_u^+} \sum_{j=1;j\sim p_i}^N -\log \sigma (\hat{y}_{ui} - \hat{y}_{uj})  \\
& =  \E_u \E_{i\sim p_u^+} \sum_{j=1; j\sim p_i}^N \log (1 + exp(\hat{y}_{uj} - \hat{y}_{ui})) 
\end{split}
\end{equation}

\noindent{\bf Softmax Loss:} 
In recommendation models, a softmax function is often utilized to transform the similarity measure into a probability measure~\cite{youtube,cl-rec}: $
p(i|u) = \frac{\exp(\hat{y}_{ui})}{\sum_{j \in \mathcal{I}} \exp(\hat{y}_{uj})}$. Given this, the maximal likelihood estimation (MLE) can be utilized to model the fit of the data through the likelihood 
($\Pi_{u\in \mathcal{U}, i \in \MI^+_u} p(i|u)$), and the negative log-likelihood serves as the loss function: 
\begin{equation}\label{eq:softmax}
\mathcal{L}_{soft} = - \E_u  \log \sum_{i \in \mathcal \MI^+_u } \frac{exp(\hat{y}_{ui})}{\sum\limits_{j \in \mathcal{I}} exp(\hat{y}_{uj})}
\end{equation} 

Note that in the earlier research, the sigmoid function has also been adopted for transforming similarity measure to probability: $p(i|u) =\sigma(\hat{y}_{ui})$. However, it has shown to be less effective ~\cite{ncfmf}. Additionally, the loss function is characterized as binary cross-entropy, wherein the negative sample component is dismissed, 
being categorized as $0$. Finally, the main challenge here is that the denominator sums over all possible items over $\mathcal{I}$, which is typically infeasible in practice and thus requires approximation. Historically, the sampled softmax approaches typically utilized the important sampling for estimation~\cite{sample-softmax}: 
\begin{equation}
    \mathcal{L}_{ssoft}= - \E_u \log \sum_{i \in \mathcal \MI^+_u } \frac{exp(\hat{y}_{ui})}{exp(\hat{y}_{ui}) + \sum\limits_{j=1; j \sim p^-_u}^N exp(\hat{y}_{uj})}
\end{equation}
where, $p_u^-$ is a predefined negative sampling distribution and often is implemented as $p$, which is proportional to the item popularity. It has been shown that sampled softmax performs better than NCE ~\cite{nce} and negative sampling ones ~\cite{bpr,sampling_strategy} when the item number is large~\cite{cl-rec}. 

\noindent{\bf Contrastive Learning Loss (InfoNCE):} 
The key idea of contrastive learning is to contrast semantically similar (positive) and dissimilar (negative) pairs of data points, thus pulling the similar points closer and pushing the dissimilar points apart. There are several contrastive learning losses have been proposed~\cite{SimCLR,infonce,unbiased}, and among them, InfoNCE loss is one of the most well-known and has been adopted in the recommendation as follows: 

\begin{equation}\label{eq:infonce_0}
   \mathcal{L}_{info}= - \E_u \sum\limits_{i \in I_u^+} \log \frac{exp(\hat{y}_{ui})}{exp(\hat{y}_{ui}) + \sum\limits_{j=1; j \sim p_u^{-}}^N  exp(\hat{y}_{uj})}
\end{equation}

where $p_u^+$ ($p_u^-$) is the positive (negative) item distribution for user $u$. 
In fact, this loss does not actually approximate the softmax probability measure $p(i|u)$ while it corresponds to the probability that {\em item $i$ is the positive item, and the remaining points are noise}, and $exp(\hat{y}_{ui})$ measures the density ration ($exp(\hat{y}_{ui}) \varpropto p(i|u)/p(i)$) ~\cite{infonce}. 
Again, the overall InfoNCE loss is the cross-entropy of identifying the positive items correctly, and it also shows maximize the mutual information between user $u$ and item $i$ ($I(i,u)$) and minimize $I(j,u)$ for item $j$ and $u$ being unrelated \citep{infonce}. 

In practice, as the true negative labels (so does all true positive labels) are not available,  negative items $j$ are typically drawn uniformly (or proportional to their popularity) from the entire item set $\mathcal{I}$ or $\mathcal{I}\backslash \MI^+_u$, i.e, $j \sim p_u$: 

\begin{equation}\label{eq:infonce_1}
   \widetilde{\mathcal{L}}_{info}= - \E_u \sum\limits_{i \in I_u^+} \log \frac{exp(\hat{y}_{ui})}{exp(\hat{y}_{ui}) + \sum\limits_{j=1; j \sim p_u}^N  exp(\hat{y}_{uj})}
\end{equation}

Note that this loss $\widetilde{\mathcal{L}}_{info}$ in practice  is sometimes treated as a softmax estimation, despite is not a proper estimator for the softmax loss~\cite{infonce,youtube}. 



\section{Decoupled Contrastive Recommendation Loss}

In this section, we begin by extending InfoNCE to a general format which we named InfoNCE+. Empirical study reveals that the selection of hyperparameter would lead to its collapse to the latest decoupled contrastive learning loss (DCL) \cite{decoupling@eccv}. As a natural extension, we delve into an exploration of DCL through the analytical lens of the Mutual Information Neural Estimator (MINE) \cite{MINE}, and further simplify InfoNCE+ as MINE+. To the best of our knowledge, this is the first time that these contrastive learning losses are introduced to the recommendation setting. 

\subsection{Generalized Contrastive Learning: InfoNCE+ }
We first consider a generalized InfoNCE loss, referred to as InfoNCE+ (\cref{eq:infonce+}), for the recommendation setting. By introducing coefficients (hyperparameters) on the denominator and inside positive-negative trade-off, InfoNCE+ enables the search of a larger space of recommendation models (with negligible alteration in model complexity), leading to potentially better model fitting and performance lift: 

\begin{small}
\begin{equation}\label{eq:infonce+}
    \begin{split}
       &\widetilde{\mathcal{L}}_{info+}=-\E_u\sum_{i\in I_u^+}(\hat{y}_{ui} - 
       \lambda \cdot \log \mathcal{N}_0)\\
       &\mathcal{N}_0 = \epsilon \cdot \exp(\hat{y}_{ui}) + \sum_{j=1; j \sim p_u}^N  exp(\hat{y}_{uj})
    \end{split}
\end{equation}
\end{small}

\begin{figure}
    \centering
    \includegraphics[width=0.8\linewidth]{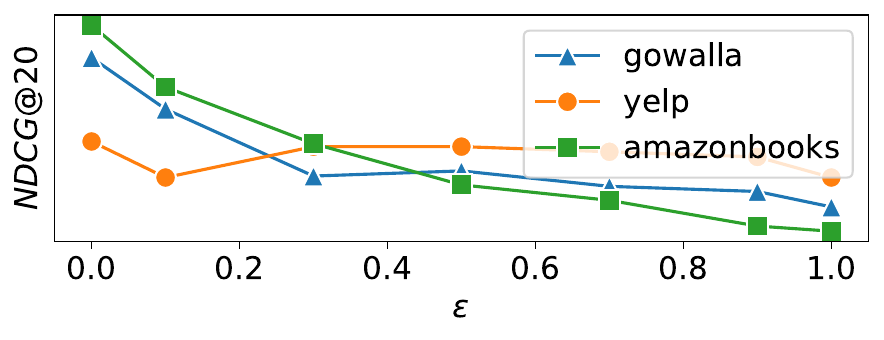}
    \vspace{-5pt}
    \caption{Empirical results on $\epsilon$ tuning. ($\lambda$ is fixed according to performance, typically around 1.1.)}
    \label{fig:pos_neg}
\end{figure}

Specifically, compared to widely-used InfoNCE (\cref{eq:infonce_1}), InfoNCE+ (\cref{eq:infonce+}) has two additional predefined coefficients $\lambda$ and $\epsilon$.  $\lambda$ controls weights of the noise in contrastive learning which has been overlooked in past studies. 
The coefficient $\epsilon$ balances the positive-negative contributions in the "noise". Surprisingly, while InfoNCE and softmax would set this value as $1$, our experimental results (in \cref{fig:pos_neg} right) reveal that when $\epsilon=0$, the recommendation models achieve peak performance. 
Interestingly, this empirical discovery is consistent with the latest decoupled contrastive learning ~\cite{decoupling@eccv}, where they omit the positive term as well. 
Finally, by default, InfoNCE chooses $\lambda$ to be 1. In our empirical studies (in \cref{fig:pos_neg} left), we found that it is typically slightly greater than the default choice. For $\epsilon=0$, it is typically around $1.1$.  This divergence explains the performance boost of the proposed InfoNCE+ in a way that emphasizes contrasted noise during optimization.

\noindent{\bf Relationship to Decoupled Contrastive Learning (DCL):}
Contrastive learning fundamentally operates by treating two augmented "views" of an identical image or representation as positives to be drawn nearer, while considering all others as negatives to be distanced. By removing \textit{negative-positive-coupling} (NPC) effect in the prevalent InfoNCE loss, \cite{decoupling@eccv} proposed Decoupled Contrastive Learning (DCL) loss which discards the positive term from the denominator and proved to improve the learning efficiency significantly. Formally, we write DCL as follows in  the recommendation setting:
\begin{equation}\label{eq:dcl}
    \begin{split}
        \widetilde{\mathcal{L}}_{dcl}=-\E_u\sum\limits_{i\in I_u^+} \Big(\hat{y}_{ui} - \log \sum\limits_{j=1;j\sim p_u^-}^N exp(\hat{y}_{uj})\Big)
    \end{split}
\end{equation}
As discussed above, this formula coincides with \cref{eq:infonce+} when $\lambda =1$ and $\epsilon=0$. Intuitively, removing the positive from the denominator gives a larger weight to the hard positives and tends
to increase the representation quality. Interestingly, this loss can be further explained from the perspective of mutual information~\cite{MINE}.

\noindent{\bf Relationship to Mutual Information Neural Estimator (MINE)}
Here, we first introduce a Mutual Information Neural Estimator (MINE) loss and show it is the equivalence of decoupling. The MINE loss is introduced in ~\cite{MINE}, which can directly measure the mutual information between user and item: 
 \begin{equation}
    \widehat{I(u,i)}=sup_{(v_u;v_i)} \E_{p_{u,i}}[\hat{y}_{ui}] - \log \E_{p_u \otimes p_i}[e^{\hat{y}_{ui}}]
\end{equation}
Here, the positive $\hat{y}_{ui}$ items sampled $n$ times according to the joint user-item distribution $p_{u,i}$. The negative items are sampled according to the empirical marginal  item distribution of $p_i$, which are proportional to their popularity (or in practice, sampled uniformly).
A simple adaption of MINE to the recommendation setting can be formalized as:

\begin{equation}\label{eq:mine}
\begin{split}
    \mathcal{L}_{mine}& =- \E_u \E_{i\sim p_u^+} \Bigg[\hat{y}_{ui}- \log \E_{j \sim p_i} [\exp \hat{y}_{uj}]\Bigg] 
\end{split}
\end{equation}



This is the same as \cref{eq:dcl} (the factor of expectation will not affect the optimization). And to the best of our knowledge, this work is the first to study and apply
MINE \cite{MINE} to the recommendation setting.

\subsection{Simplified Contrastive Learning: MINE+}
Ultimately, with the support of our empirical study and the theoretical foundation of decoupled contrastive learning as well as MINE, we propose the following simplified InfoNCE+ (\cref{eq:infonce+}, which is referred to as MINE+:  

\begin{equation}\label{eq:mine+}
\boxed{
\begin{split}
    \widetilde{\mathcal{L}}_{mine+}=- \E_u \sum\limits_{i\sim p_u^+} & \Bigg[{\hat{y}_{ui}}- \lambda \log \sum_{j=1; j \sim p_i}^N \exp  {\hat{y}_{uj}}\Bigg] 
\end{split}
}
\end{equation}
In addition, $\hat{y}_{ui}$ can be adopted as cosine similarity between user and item latent vectors, i.e., cosine ($<v_u/||v_u||, v_i/||v_i||>$),  then $\hat{y}_{ui}$ is normalized by a temperature, such as $<v_u/||v_u||, v_i/||v_i||>/t$. Note that such treatment of using temperature is a common practice for the contrastive learning~\cite{SimCLR}.  Our experimental results also confirm such treat can help boost the performance for recommendation as well. 

\subsection{Lower Bound Analysis}


In this subsection, we theoretically discuss some common lower bounds for InfoNCE and MINE (aka DCL), BPR as well as their relationship. 

Given \Cref{eq:mine}, we can easily observe (based on Jensen's inequality): 
\begin{small}
\begin{equation}\label{eq:loose_bound}
\begin{split}
    &\widetilde{\mathcal{L}}_{info} = - \E_u \E_{i \sim p_u^+} \log \frac{exp(\hat{y}_{ui})}{exp(\hat{y}_{ui}) + \sum_{j=1; j \sim p_u}^N  exp(\hat{y}_{uj})} \geq \\
    & \widetilde{\mathcal{L}}_{mine} =  
    \E_u \E_{i\sim p_u^+} \log (\sum_{j=1; j\sim p_i}^N exp(\hat{y}_{uj} - \hat{y}_{ui})) \approx \\  
    & \E_u \E_{i\sim p_u^+} \log N \E_{j\sim p_i} exp(\hat{y}_{uj} - \hat{y}_{ui})) \geq  \\
    & \E_u \E_{i\sim p_u^+} \E_{j\sim p_i} \bigl( \hat{y}_{uj} - \hat{y}_{ui}  \bigr) + \log N 
    \geq \E_u \E_{i\sim p_u^+} \E_{j\sim p_i} \bigl( \hat{y}_{uj} - \hat{y}_{ui}  \bigr)
\end{split}
\end{equation}
\end{small}
Alternative, by using LogSumExp lower-bound~\cite{unbiased}, we also observe:  
\begin{small}
\begin{equation}\label{eq:tight_bound} 
\begin{split}
    & \E_u \E_{i\sim p_u^+} \max_{j=1; j\sim p_i} ^N \bigl( 0, \hat{y}_{uj} - \hat{y}_{ui}  \bigr)  + \log (N+1) \geq \\
     & \widetilde{\mathcal{L}}_{info} \geq 
     \widetilde{\mathcal{L}}_{mine} =  
    \E_u \E_{i\sim p_u^+} \log (\sum_{j=1; j\sim p_i}^N exp(\hat{y}_{uj} - \hat{y}_{ui})) \geq \\ 
    &\E_u \E_{i\sim p_u^+} \max_{j=1; j\sim p_i} ^N \bigl( \hat{y}_{uj} - \hat{y}_{ui}  \bigr)
\end{split}
\end{equation}
\end{small}
Note that the second lower bound is tighter than the first one. And it is obvious to observe that the looser bound \cref{eq:loose_bound} is also shared by the well-known pairwise BPR loss (\cref{eq:bpr}), which aims to maximize the average difference between a positive item score ($\hat{y}_{ui}$) and a negative item score ($\hat{y}_{uj}$).

However, the tighter lower bound of BPR diverges from the one shared by InfoNCE and MINE loss (\cref{eq:tight_bound}) using the LogSumExp lower bound: 
\begin{small}
\begin{equation}\label{eq:bound_bpr}
\begin{split}
&\widetilde{\mathcal{L}}_{bpr} =  \E_u \E_{i\sim p_u^+} \sum_{j=1; j\sim p_i}^N \log (1 + exp(\hat{y}_{uj} - \hat{y}_{ui})) \geq \\
& \E_u \E_{i\sim p_u^+} \sum_{j=1; j\sim p_i}^N 
max(0, \hat{y}_{uj} - \hat{y}_{ui}) 
\varpropto \\ & \E_u \E_{i\sim p_u^+} \E_{j\sim p_i}
max(0, \hat{y}_{uj} - \hat{y}_{ui})
\end{split}
\end{equation}
\end{small}
Note that the bound in \cref{eq:bound_bpr} only sightly improves over the one in \cref{eq:loose_bound} and cannot reflect the tight bounds (\cref{eq:tight_bound}) shared by InfoNCE and MINE, which aims to minimize the largest gain of a negative item (random) over a positive item ($\max_{j=1; j\sim p_i} ^N \bigl( \hat{y}_{uj} - \hat{y}_{ui}  \bigr)$).

This bound (\cref{eq:tight_bound}) provides a deeper understanding of the contrastive learning loss, which aims to minimize the largest gain of a negative item (random) over a positive item ($\max_{j=1; j\sim p_i} ^N \bigl( \hat{y}_{uj} - \hat{y}_{ui}  \bigr)$).

\section{Debiased Contrastive Recommendation Loss}

In ~\cite{debiased}, a novel debiased estimator is proposed while only accessing positive items and unlabeled training data. Below, we demonstrate how this approach can be adopted in the recommendation setting. 

Given user $u$, let us consider the implicit (binary) preference probability (class distribution) of $u$: $\rho_u(c)$, where $\rho_u(c=+)=\tau^+_u$ represents the probability of user $u$ likes an item, and $\rho_u(c=-)=\tau^-_u=1-\tau^+_u$ corresponds to being not interested. In  the recommendation set, we may consider $\rho_u(+)$ as the fraction of positive items, i.e., those top $K$ items, and  $\rho_u(-)$ are the fraction of true negative items.  Let the item class joint distribution be $p_u (i,c)=p_u(i|c)\rho_u(c)$. 
Then $p^+_u(i)=p_u(i|c=+)$ is the probability of observing $i$ as a positive item for user $u$ and $p_u^-(j)=p_u(j|c=-)$ the probability of a negative example. Given this, the distribution of an individual item of user $u$ is:
\begin{small}
\begin{equation}
    \begin{split}
        p_u(i) = \tau^+_u \cdot p^+_u(i) + \tau^-_u \cdot p_u^-(i)
    \end{split}
\end{equation}
\end{small}

Assume the known positive items $\mathcal{I}_u^+$ are being uniformly sampled according to $p^+_u$. For $\tau^+_u$, we can count all the existing positive items ($r_{ui}=1$), and the unknown top $K$ items as being positive, thus, $\tau^+_u=(|\MI_{u}^+|+K)/|\mathcal{I}|$. Alternatively, we can also assume the number of total positive items for $u$ is proportional to the number of existing (known) positive items, i.e, $\tau^+_u=(1+\alpha) |\MI_{u}^+|/|\mathcal{I}|$, where $\alpha\geq 0$. As mentioned earlier, the typical assumption of $p_u$ (the probability distribution of selection of an item is also uniform). 
Then, we can represent $p^-_{u}$ using the total and positive sampling as: $p^-_{u}(i)=1/\tau^-_u (p_u(i)-\tau^+_u \cdot p^+_u(i))$. Upon these assumptions, let us consider the following ideal (debiased) InfoNCE loss: 
\begin{small}
\begin{equation}
\begin{split}
    \overline{\mathcal{L}}_{Info}&=- \E_{u} \E_{i \sim p_u^+}\Bigg[\log\frac{\exp(\hat{y}_{ui})}{\exp(\hat{y}_{ui}) + \lambda_n \E_{j\sim p_u^-} \exp({\hat{y}_{uj}}) }
    \Bigg]  \\
    &=- \E_{u} \E_{i\sim p_u^+}\Bigg[\log\frac{\exp(\hat{y}_{ui})}{\exp(\hat{y}_{ui}) + \lambda_n \frac{1}{\tau^-}T }
    \Bigg]\\
    T &= \Big( 
    \E_{t\sim p_u}[\exp(\hat{y}_{ut})] - \tau^+ \E_{j\sim p^+_u}[\exp(\hat{y}_{uj})]\Big)
\end{split}
\end{equation}
\end{small}
Note that when $\lambda_n=N$, the typical InfoNCE loss $\mathcal{L}_{Info}$ (\cref{eq:infonce_0}) converges into $\overline{\mathcal{L}}_{Info}$ when $N \rightarrow \infty$.
Now, we can use $N$ samples from $p_u$ (not negative samples), an $M$ positive samples from $p_u^+$ to empirically estimate $\overline{\mathcal{L}}_{Info}$ following ~\cite{debiased}: 
\begin{small}
\begin{equation}\label{eq:debiased_infonce}
\begin{split}
    \mathcal{L}^{Debiased}_{Info}=-\E_{u} \E_{ \substack{i \sim p_u^+;\\ j \sim p_u; \\ k \sim p_u^+} }\Bigg[\log\frac{\exp(\hat{y}_{ui})}{\exp(\hat{y}_{ui}) + \lambda_n f(\{j\}_{1}^N, \{k\}_{1}^M)}
    \Bigg] 
\end{split}
\end{equation}
\end{small}    

$f(\{ j\}_{1}^N, \{k\}_{1}^M)= 
    \max \Bigl\{ \frac{1}{\tau_u^-} \Bigl( \frac{1}{N} \sum_{j=1;j\sim p_u}^N exp(f_{uj})- \tau_u^+\frac{1}{M} \sum_{k=1;k\sim p_u^+}^M exp(f_{uk}) \Bigr), e^{1/t}  \Bigr\}$. Here $f$ is constrained to be greater than its theoretical minimum $e^{-1/t}$ to prevent the negative number inside $\log$.
We will experimentally study this debiased InfoNCE in a later section.

\subsection{Debiased CCL}
Following the analysis of debiased contrastive learning, clearly, some of the positive items are unlabeled, and making them closer to $0$ is not the ideal option. In fact, if the optimization indeed reaches the targeted score, we actually do not learn any useful recommendations. A natural question is whether we can improve other contrastive learning loss (for example, cosine contrastive loss (CCL) \cite{simplex}) following the debiased contrastive learning paradigm ~\cite{debiased}?

The cosine contrastive loss (CCL) is a recently proposed loss and has shown to be very effective for recommendation~\cite{simplex}.
The original Contrastive Cosine Loss (CCL) \citep{simplex} can be written as:
\begin{small}
\begin{equation}
    \begin{split}
        \mathcal{L}_{ccl}=\E_{u}
        \Bigg(\sum\limits_{i\in \mathcal{I}^+_u} (1-\hat{y}_{ui}) + \frac{w}{N}\sum\limits_{j\in\mathcal{I}}^N ReLU(\hat{y}_{uj}-\epsilon_0)\Bigg)
    \end{split}
\end{equation}
\end{small}

where $\hat{y}_{uj}$ is cosine similarity, $ReLU(x) = max(0, x)$ is the activation function, $w$ is the negative weight, $N$ is the number of negative samples and $\epsilon_0$ is margin.


\bdefin{\bf (Ideal CCL Loss)}
The ideal CCL loss function for recommendation models is the expected single pointwise loss for all positive items (sampled from $\rho_u(+)=
\tau^+$) and for negative items (sampled from $\rho_u(-)=
\tau^-$):  
\begin{small}
\begin{equation}\label{eq:ideal_ccl}
    \overline{\mathcal{L}}_{ccl} = \E_{u} \Bigg(\tau_u^+\E_{i\sim p_u^+}(1-\hat{y}_{ui}) +
    \lambda_n\cdot \tau_u^-\E_{j\sim p_u^-}[ReLU(\hat{y}_{uj}-\epsilon_0)]\Bigg)
\end{equation}
\end{small}
\edefin

Why this is ideal? When the optimization reaches the potential minimal, we have all positive items $i$ and all the negative items $j$ reaching all the minimum of individual loss ($l^+$ and $l^-$).  
Here, $\lambda_n$ help adjusts the balance between the positive and negative losses.
However, since $p_u^+$ and $p_u^-$ are unknown, we utilize the same debiase approach as InfoNCE ~\cite{infonce}.
\bdefin {\bf (Debiased CCL loss)}
\begin{small}
\begin{equation}\label{eq:debiased_ccl}
\boxed{
    \begin{split}
    &\mathcal{L}^{debiased}_{CCL} = \E_{u} \Bigg(\tau_u^+\E_{i\sim p_u^+}(1-\hat{y}_{ui}) \\  
    &+  \lambda_n\Big( 
    \E_{j\sim p_u}[ReLU(\hat{y}_{uj}-\epsilon)] - \tau_u^+ \E_{k\sim p^+_u}[ReLU(\hat{y}_{uk}-\epsilon_0)]\Big)\Bigg) \\
    \end{split}
}
\end{equation}
\end{small}
\edefin

In practice, for the computation of expectation, we sample $N$ positive items and $M$ negative items to obtain an empirical loss.

\subsection{iALS and EASE are debiased}
Besides the typical InfoNCE as well as cosine contrastive learning loss, another most commonly used loss is Mean-Square-Error (MSE), as well as the well-known quadratic linear models - iALS \cite{ials_revisiting,ials@hu2008} and EASE \cite{ease}. We may wonder whether they are biased or not. Surprisingly, we found these linear models are naturally debiased which contributes to the following observation:

\bobserv{}\label{obs:debiased_ials_ease}
The solvers of both iALS and EASE models can absorb their debiased counterparts under existing frameworks with
reasonable conditions.
\eobserv
This observation reflects the outstanding performances of simple linear models from the debias angle, which coincide with the recent series of fundamental revisiting works \cite{ials_revisiting,RecSys19Evaluation,jin@linear}. For detailed proof and explanation please see \textbf{Theorem 1} and \textbf{Theorem 2} and their proofs in technical supplementary.

\begin{table*}[]
\caption{Performance comparison to widely used models. We highlight the top-2 best models in each column. Results of models marked with $*$ are duplicated from \citep{simplex} for consistency. For a fair comparison, MF-based models marked with $**$ are reproduced by ourselves with thorough parameter searching.}
\label{tab:main}
\begin{small}
\begin{tabular}{|ccccccc|}
\hline
\multicolumn{1}{|c|}{\multirow{2}{*}{Model}} & \multicolumn{2}{c|}{Yelp}                                               & \multicolumn{2}{c|}{Gowalla}                                              & \multicolumn{2}{c|}{Amazon-Books}                  \\ \cline{2-7} 
\multicolumn{1}{|c|}{}                       & \multicolumn{1}{c|}{Recall@20}     & \multicolumn{1}{c|}{NDCG@20}       & \multicolumn{1}{c|}{Recall@20}      & \multicolumn{1}{c|}{NDCG@20}        & \multicolumn{1}{c|}{Recall@20}     & NDCG@20       \\ \hline
\multicolumn{7}{|c|}{Deep Learning Based}                                                                                                                                                                                                               \\ \hline
\multicolumn{1}{|c|}{YouTubeNet* \citep{youtube}}            & \multicolumn{1}{c|}{6.86}          & \multicolumn{1}{c|}{{5.67}} & \multicolumn{1}{c|}{17.54}          & \multicolumn{1}{c|}{14.73}          & \multicolumn{1}{c|}{5.02}          & 3.88          \\ \hline
\multicolumn{1}{|c|}{NeuMF* \citep{ncf}}                 & \multicolumn{1}{c|}{4.51}          & \multicolumn{1}{c|}{3.63}          & \multicolumn{1}{c|}{13.99}          & \multicolumn{1}{c|}{12.12}          & \multicolumn{1}{c|}{2.58}          & 2.00             \\ \hline
\multicolumn{1}{|c|}{CML* \citep{cml}}                   & \multicolumn{1}{c|}{6.22}          & \multicolumn{1}{c|}{5.36}          & \multicolumn{1}{c|}{16.7}           & \multicolumn{1}{c|}{12.92}          & \multicolumn{1}{c|}{{5.22}} & 4.28          \\ \hline
\multicolumn{1}{|c|}{MultiVAE* \citep{multi-vae}}              & \multicolumn{1}{c|}{5.84}          & \multicolumn{1}{c|}{4.50}           & \multicolumn{1}{c|}{16.41}          & \multicolumn{1}{c|}{13.35}          & \multicolumn{1}{c|}{4.07}          & 3.15          \\ \hline
\multicolumn{1}{|c|}{LightGCN* \citep{lightgcn}}              & \multicolumn{1}{c|}{6.49}          & \multicolumn{1}{c|}{5.30}           & \multicolumn{1}{c|}{{18.3}}  & \multicolumn{1}{c|}{\textbf{15.54}} & \multicolumn{1}{c|}{4.11}          & 3.15          \\ \hline
\multicolumn{1}{|c|}{NGCF* \citep{ngcf}}                  & \multicolumn{1}{c|}{5.79}          & \multicolumn{1}{c|}{4.77}          & \multicolumn{1}{c|}{15.7}           & \multicolumn{1}{c|}{13.27}          & \multicolumn{1}{c|}{3.44}          & 2.63          \\ \hline
\multicolumn{1}{|c|}{GAT* \citep{gat}}                   & \multicolumn{1}{c|}{5.43}          & \multicolumn{1}{c|}{4.32}          & \multicolumn{1}{c|}{14.01}          & \multicolumn{1}{c|}{12.36}          & \multicolumn{1}{c|}{3.26}          & 2.35          \\ \hline
\multicolumn{1}{|c|}{PinSage* \citep{pinsage}}               & \multicolumn{1}{c|}{4.71}          & \multicolumn{1}{c|}{3.93}          & \multicolumn{1}{c|}{13.80}           & \multicolumn{1}{c|}{11.96}          & \multicolumn{1}{c|}{2.82}          & 2.19          \\ \hline
\multicolumn{1}{|c|}{SGL-ED* \citep{sgl-ed}}                & \multicolumn{1}{c|}{6.75}          & \multicolumn{1}{c|}{5.55}          & \multicolumn{1}{c|}{-}              & \multicolumn{1}{c|}{-}              & \multicolumn{1}{c|}{4.78}          & 3.79          \\ \hline
\multicolumn{7}{|c|}{MF based}                                                                                   \\ \hline
\multicolumn{1}{|c|}{iALS** \citep{ials_revisiting}}                & \multicolumn{1}{c|}{6.06}          & \multicolumn{1}{c|}{5.02}          & \multicolumn{1}{c|}{13.88}          & \multicolumn{1}{c|}{12.24}          & \multicolumn{1}{c|}{2.79}          & 2.25        \\
\hline
\multicolumn{1}{|c|}{MF-BPR** \citep{bpr}}                & \multicolumn{1}{c|}{5.63}          & \multicolumn{1}{c|}{4.60}          & \multicolumn{1}{c|}{15.90}          & \multicolumn{1}{c|}{13.62}          & \multicolumn{1}{c|}{3.43}          & 2.65         \\ \hline
\multicolumn{1}{|c|}{MF-CCL** \citep{simplex}}               & \multicolumn{1}{c|}{{6.91}} & \multicolumn{1}{c|}{{5.67}} & \multicolumn{1}{c|}{{18.17}} & \multicolumn{1}{c|}{14.61}          & \multicolumn{1}{c|}{{5.27}} & \textbf{4.22} \\ \hline
\multicolumn{1}{|c|}{\textbf{MF-CCL-debiased (ours)}} & \multicolumn{1}{c|}{\textbf{6.98}} & \multicolumn{1}{c|}{\textbf{5.71}} & \multicolumn{1}{c|}{\textbf{18.42}} & \multicolumn{1}{c|}{{14.97}} & \multicolumn{1}{c|}{\textbf{5.49}} & \textbf{4.40} \\ \hline
\multicolumn{1}{|c|}{\textbf{MF-MINE+ (ours)}}        & \multicolumn{1}{c|}{\textbf{7.20}} & \multicolumn{1}{c|}{\textbf{5.92}} & \multicolumn{1}{c|}{\textbf{18.53}}          & \multicolumn{1}{c|}{\textbf{15.70}} & \multicolumn{1}{c|}{\textbf{5.31}}          & {4.14} \\ \hline
\end{tabular}
\end{small}
\end{table*}

\section{Experiment}
\label{sec:experiment}

In this section, we experimentally study the newly proposed MINE+ loss as well as the existing recommendation loss w.r.t their corresponding debiased ones. This could help better understand and compare the resemblance and discrepancy between different loss functions in the perspective of contrastive learning.  Specifically, we would like to answer the following questions:
\begin{itemize}
\item Q1. How does our proposed MINE+ objective function compare to widely used ones?
\item Q2. How does the bias introduced by the existing negative sampling impact model performance and how do our proposed debiased losses perform compared to traditional biased ones?
\item Q3. What is the effect of different hyperparameters on the contrastive learning loss functions?
\end{itemize}

\subsection{Experimental Setup}

\noindent\textbf{Datasets}
We examed three commonly used datasets, \textbf{Amazon-Books}, \textbf{Yelp2018} and \textbf{Gowalla} by a number of recent studies \citep{simplex,lightgcn,ngcf,sgl-ed}. We obtain the publicly available processed data and follow the same setting as \citep{ngcf,lightgcn,simplex}.

\noindent\textbf{Evaluation Metrics} Consistent with benchmarks \citep{ngcf,lightgcn,simplex}, we evaluate the performance by $Recall@20$ and $NDCG@20$ over all items \citep{walid@sample,Jin@AAAI21,li@acm23,Li@KDD20,dong@2023aaai}.

\noindent\textbf{Models}
This paper focuses on the study of loss functions, which is architecture agnostic. Thus, we evaluate our loss functions on top of the simplest one - Matrix Factorization (MF) in this study. By eliminating the influence of the architecture, we can compare the power of various loss functions fairly. 
Meantime, we compare the performance of the proposed methods with a few classical machine learning models as well as advanced deep learning models, including, iALS \citep{ials_revisiting}, MF-BPR\citep{bpr}, MF-CCL \citep{simplex}, YouTubeNet \citep{youtube}, NeuMF \citep{ncf}, LightGCN \citep{lightgcn}, NGCF \citep{ngcf}, CML \citep{cml}, PinSage \citep{pinsage}, GAT \citep{gat}, MultiVAE \citep{multi-vae}, SGL-ED \citep{sgl-ed}.
Those comparisons can help establish that the loss functions with basic MF can produce strong and robust baselines for the recommendation community to evaluate more advanced and sophisticated models. 

\noindent\textbf{Loss functions}
As the core of this paper, we focus on some most widely used loss functions and their debiased counterparts, including InfoNCE \citep{infonce}, MSE \citep{ials@hu2008,ials_revisiting}, and state-of-the-art CCL \citep{simplex}, as well as our new proposed MINE+ \citep{MINE,decoupling@eccv}.

\noindent\textbf{Reproducibility}
To implement reliable models, by default, we set batch size as $512$. Adam optimizer learning rate is initially set as $1e-4$ and reduced by $0.5$ on the plateau, early stopped until it arrives at $1e-6$. For the cases where negative samples get involved, we would set it as $800$ by default. We search the global regularization weight between $1e-9$ to $1$ with an increased ratio of $10$. To ensure the reproducibility, for the exact hyperparameter settings of the model, we would present it in the technical appendix and our anonymous code is also available at \url{https://anonymous.4open.science/r/AAAI-2024-Anonymous-9B33/README.md}.

\begin{table*}[]
\caption{The performance of different loss functions w.r.t their debiased ones on MF. The unit of the metric values is $\%$. We also present and highlight the relative improvement (RI) of the debiased loss function with its vanilla (biased) counterpart.}
\label{tab:debias}
\begin{small}
\begin{tabular}{|cc|cc|cc|cc|c|}
\hline
\multicolumn{2}{|c|}{\multirow{2}{*}{Loss}}                                                                                                                           & \multicolumn{2}{c|}{Yelp}                              & \multicolumn{2}{c|}{Gowalla}                             & \multicolumn{2}{c|}{Amazon-Books}                        & \multirow{2}{*}{Average RI\%} \\ \cline{3-8}
\multicolumn{2}{|c|}{}                                                                                                                                                & \multicolumn{1}{c|}{Recall@20}       & NDCG@20         & \multicolumn{1}{c|}{Recall@20}        & NDCG@20          & \multicolumn{1}{c|}{Recall@20}        & NDCG@20          &                               \\ \hline
\multicolumn{1}{|c|}{\multirow{3}{*}{CCL}}                                                   & biased                                                                 & \multicolumn{1}{c|}{6.91}            & 5.67            & \multicolumn{1}{c|}{18.17}            & 14.61            & \multicolumn{1}{c|}{5.27}             & 4.22             & -                             \\ \cline{2-9} 
\multicolumn{1}{|c|}{}                                                                       & debiased                                                               & \multicolumn{1}{c|}{6.98}            & 5.71            & \multicolumn{1}{c|}{18.42}            & 14.97            & \multicolumn{1}{c|}{5.49}             & 4.40             & -                             \\ \cline{2-9} 
\multicolumn{1}{|c|}{}                                                                       & RI\%                                                                   & \multicolumn{1}{c|}{\textbf{1.01\%}} & \textbf{0.71\%} & \multicolumn{1}{c|}{\textbf{1.38\%}}  & \textbf{2.46\%}  & \multicolumn{1}{c|}{\textbf{4.17\%}}  & \textbf{4.27\%}  & \textbf{2.33\%}               \\ \hline


\multicolumn{1}{|c|}{\multirow{3}{*}{InfoNCE}}                                               & biased                                                                 & \multicolumn{1}{c|}{6.54}            & 5.36            & \multicolumn{1}{c|}{16.45}            & 13.43            & \multicolumn{1}{c|}{4.60}             & 3.56             & -                             \\ \cline{2-9} 
\multicolumn{1}{|c|}{}                                                                       & debiased                                                               & \multicolumn{1}{c|}{6.66}            & 5.45            & \multicolumn{1}{c|}{16.57}            & 13.72            & \multicolumn{1}{c|}{4.65}             & 3.66             & -                             \\ \cline{2-9} 
\multicolumn{1}{|c|}{}                                                                       & RI\%                                                                   & \multicolumn{1}{c|}{\textbf{1.83\%}} & \textbf{1.68\%} & \multicolumn{1}{c|}{\textbf{0.73\%}}  & \textbf{2.16\%}  & \multicolumn{1}{c|}{\textbf{1.09\%}}  & \textbf{2.81\%}  & \textbf{1.72\%}               \\ \hline
\multicolumn{1}{|c|}{\multirow{4}{*}{\begin{tabular}[c]{@{}c@{}}MINE\\ (ours)\end{tabular}}} & MINE                                                                  & \multicolumn{1}{c|}{6.56}            & 5.37            & \multicolumn{1}{c|}{16.93}            & 14.28            & \multicolumn{1}{c|}{5.00}             & 3.93             & -                             \\ \cline{2-9} 
\multicolumn{1}{|c|}{}                                                                       & \begin{tabular}[c]{@{}c@{}}RI\% \\ (w.r.t biased infoNCE)\end{tabular} & \multicolumn{1}{c|}{\textbf{0.31\%}} & \textbf{0.19\%} & \multicolumn{1}{c|}{\textbf{2.92\%}}  & \textbf{6.33\%}  & \multicolumn{1}{c|}{\textbf{8.70\%}}  & \textbf{10.39\%} & \textbf{4.80\%}               \\ \cline{2-9} 
\multicolumn{1}{|c|}{}                                                                       & MINE+                                                                  & \multicolumn{1}{c|}{7.20}            & 5.92            & \multicolumn{1}{c|}{18.53}            & 15.70            & \multicolumn{1}{c|}{5.31}             & 4.14            & -                             \\ \cline{2-9} 
\multicolumn{1}{|c|}{}                                                                       & \begin{tabular}[c]{@{}c@{}}RI\% \\ (w.r.t biased infoNCE)\end{tabular} & \multicolumn{1}{c|}{\textbf{10.09\%}} & \textbf{10.45\%} & \multicolumn{1}{c|}{\textbf{12.64\%}} & \textbf{16.90\%} & \multicolumn{1}{c|}{\textbf{15.43\%}} & \textbf{16.29\%} & \textbf{13.63\%}              \\ \hline
\end{tabular}
\end{small}
\end{table*}

\subsection{Q1. Performance of MINE/MINE+}
First, we delve into the examination of the newly proposed MINE+ loss function (\cref{eq:mine+}). As mentioned before, to our best knowledge, this is the first study utilizing MINE/MINE+ aka decoupled contrastive learning loss in recommendation settings. In \cref{tab:main}, we can observe MF-MINE+, as well as debiased MF-CCL, perform best or second best in most cases, affirming the superiority of the MINE+ loss and debiased CCL. Then, We compare MINE/MINE+ with other loss functions (CCL, InfoNCE), all set MF as the backbone, as displayed in (\cref{tab:debias},. 
Our results show the surprising effectiveness of MINE-type losses.
In particular, the basic MINE is shown close to $4.8\%$ average lifts over the (biased) InfoNCE loss, a rather stand loss in recommendation models. MINE+ even demonstrates superior performance compared to the state-of-the-art CCL loss function. Furthermore, MINE+ consistently achieves better results than InfoNCE with a maximum improvement of 16\% (with an average improvement over $14\%$. 

These findings highlight the potential of MINE as an effective objective function for recommendation models and systems.  
Noting that our MINE+ loss is model agnostic, which can be applied to any representation-based model. We leave the study of combining MINE.MINE+ models with more advanced (neural) architecture in future studies.


\begin{figure}
    \centering
    \includegraphics[width=0.85\linewidth]{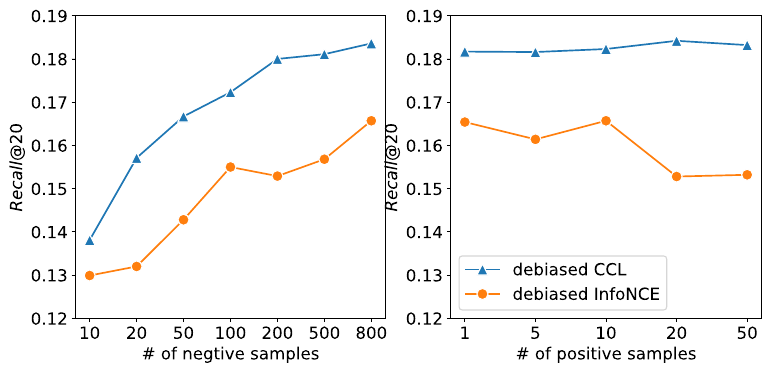}
    \vspace{-8pt}
    \caption{Effect of number of samples on $Gowalla$}
    \label{fig:hyper_pos_neg}
    \vspace{-5pt}
\end{figure}

\subsection{Q2. (Biased) Loss vs Debiased Loss}
Here, we seek to examine the negative impact of bias introduced by negative sampling and assess the effectiveness of our proposed debiased loss functions. We utilize Matrix Factorization (MF) as the backbone and compare the performance of the original (biased) version and the debiased version for the loss functions of $CCL$ \citep{simplex} and $InfoNCE$ \citep{infonce}. \cref{tab:debias} shows the biased loss vs debiased loss (on top of the MF) results. 

The results show that the debiased methods consistently outperform the biased ones for the CCL and InfoNCE loss functions in all cases. In particular, the debiased method exhibits remarkable improvements of up to over $4\%$ for the CCL loss function on the Amazon-Books dataset. This highlights the adverse effect of bias in recommendation quality and demonstrates the success of applying contrastive learning based debiase methods in addressing this issue.
Finally, we note that the performance gained by the debiased losses is also consistent with the results over other machine learning tasks as shown in ~\cite{debiased}.  

\subsection{Q3. Hyperparameter analysis}
\begin{figure}
\vspace{-8pt}
    \centering
    \includegraphics[width=0.9\linewidth]{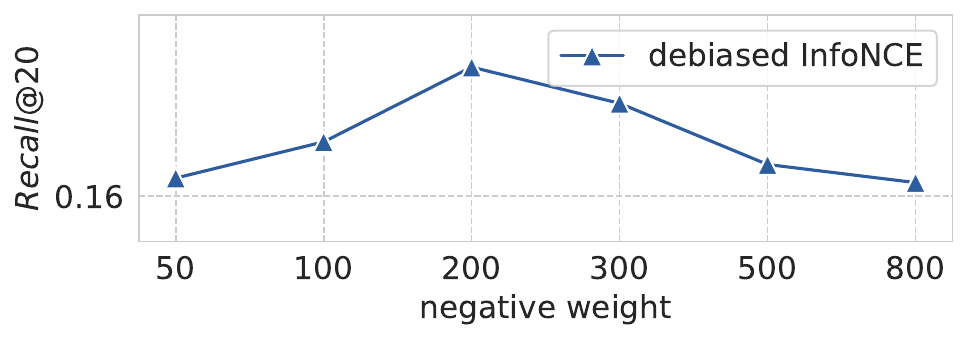}
    \vspace{-8pt}
    \caption{Effect of negative weight $\lambda_0$ on debiased InfoNCE loss $Gowalla$}
    \label{fig:hyper_lambda}
    \vspace{-5pt}
\end{figure}

\begin{figure}
    \centering
    \includegraphics[width=0.8\linewidth]{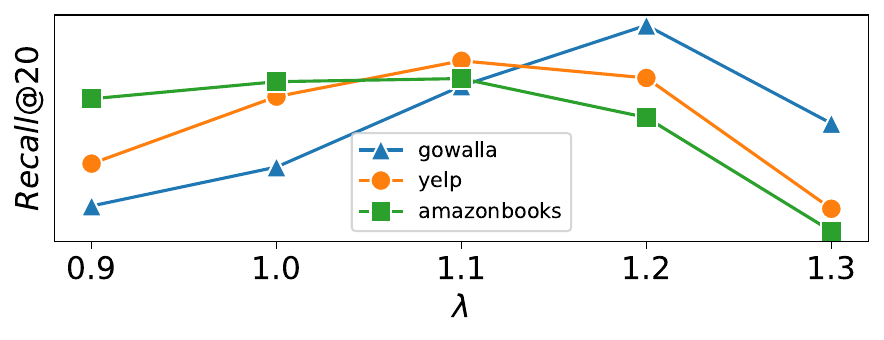}
    \vspace{-8pt}
    \caption{Effect of noise weight $\lambda$ in MINE+ loss on $Gowalla$}
    \label{fig:lamd_mine+}
\end{figure}

First, we analyze the impacts of hyperparameters in the debiased framework (\cref{eq:debiased_infonce,eq:debiased_ccl}): negative weight $\lambda_n$, number of negative samples, and number of positive samples as shown in \cref{fig:hyper_pos_neg,fig:hyper_lambda}. The left side of \cref{fig:hyper_pos_neg} reveals that increasing the number of negative samples leads to better performance in models overall. The right side of \cref{fig:hyper_pos_neg} shows that the number of positive samples in the debiased formula has a minor impact on performance compared to negative samples. In \cref{fig:hyper_lambda}, we illustrate how the negative weight $\lambda_n$ affects the performance of the debiased InfoNCE (\cref{eq:debiased_infonce}). In general, we observe the $\cap$ shaped curve for quality metric ($Recall@20$ here) for the parameters. In \cref{fig:lamd_mine+}, we investigate how noise weight $\lambda$ in \cref{eq:mine+} can affect the performance of the $MINE+$ loss function. In general, it would achieve optimal around $1.1\sim 1.2$ for all datasets. 


\section{conclusion}
\label{sec:conclusion}
In this paper, we conduct a comprehensive analysis of recommendation loss functions from the perspective of contrastive learning. 
Two novel recommendation losses MINE+ and Debiased CCL are developed through investigation into and integration with the contrastive learning losses.  Both iALS and EASE are certified to be inherently debiased. 
The empirical experimental results demonstrate the debiased losses and new information losses outperform the existing (biased) ones. 
In the future, we would like to investigate how effective does the loss functions work with more sophisticated neural architectures and seek additional theoretical evidence on why MINE+ performs better than other losses, such as SoftMax.



\section*{Acknowledgement}

This research was partially funded by the National Science Foundation through grants IIS-2142675, IIS-2142681, and III-2008557. Additional support came from a collaborative research agreement between Kent State University and iLambda Inc.
\bibliography{aaai24}

\begin{thebibliography}{63}
\providecommand{\natexlab}[1]{#1}

\bibitem[{Aggarwal(2016)}]{charubook}
Aggarwal, C.~C. 2016.
\newblock \emph{Recommender Systems: The Textbook}.
\newblock Springer Publishing Company, Incorporated, 1st edition.
\newblock ISBN 3319296574.

\bibitem[{Barbano et~al.(2023)Barbano, Dufumier, Tartaglione, Grangetto, and Gori}]{unbiased}
Barbano, C.~A.; Dufumier, B.; Tartaglione, E.; Grangetto, M.; and Gori, P. 2023.
\newblock Unbiased Supervised Contrastive Learning.
\newblock In \emph{International Conference on Learning Representations}.

\bibitem[{Belghazi et~al.(2018)Belghazi, Baratin, Rajeshwar, Ozair, Bengio, Courville, and Hjelm}]{MINE}
Belghazi, M.~I.; Baratin, A.; Rajeshwar, S.; Ozair, S.; Bengio, Y.; Courville, A.; and Hjelm, D. 2018.
\newblock Mutual information neural estimation.
\newblock In \emph{International conference on machine learning}, 531--540. PMLR.

\bibitem[{Chapelle and Zhang(2009)}]{dbcl}
Chapelle, O.; and Zhang, Y. 2009.
\newblock A Dynamic Bayesian Network Click Model for Web Search Ranking.
\newblock In \emph{Proceedings of the 18th International Conference on World Wide Web}, WWW '09, 1–10. New York, NY, USA: Association for Computing Machinery.
\newblock ISBN 9781605584874.

\bibitem[{Chen et~al.(2022{\natexlab{a}})Chen, Ma, Zhang, Wang, Liu, and Ma}]{negative_sampling_review}
Chen, C.; Ma, W.; Zhang, M.; Wang, C.; Liu, Y.; and Ma, S. 2022{\natexlab{a}}.
\newblock Revisiting Negative Sampling VS. Non-Sampling in Implicit Recommendation.
\newblock \emph{ACM Trans. Inf. Syst.}

\bibitem[{Chen et~al.(2022{\natexlab{b}})Chen, Dong, Wang, Feng, Wang, and He†}]{bias-debias}
Chen, J.; Dong, H.; Wang, X.; Feng, F.; Wang, M.; and He†, X. 2022{\natexlab{b}}.
\newblock Bias and Debias in Recommender System: A Survey and Future Directions.
\newblock \emph{ACM Trans. Inf. Syst.}
\newblock Just Accepted.

\bibitem[{Chen et~al.(2020)Chen, Kornblith, Norouzi, and Hinton}]{SimCLR}
Chen, T.; Kornblith, S.; Norouzi, M.; and Hinton, G. 2020.
\newblock A simple framework for contrastive learning of visual representations.
\newblock In \emph{International conference on machine learning}, 1597--1607. PMLR.

\bibitem[{Chen et~al.(2017)Chen, Sun, Shi, and Hong}]{sampling_strategy}
Chen, T.; Sun, Y.; Shi, Y.; and Hong, L. 2017.
\newblock On Sampling Strategies for Neural Network-Based Collaborative Filtering.
\newblock In \emph{Proceedings of the 23rd ACM SIGKDD International Conference on Knowledge Discovery and Data Mining}, KDD '17, 767–776. New York, NY, USA: Association for Computing Machinery.
\newblock ISBN 9781450348874.

\bibitem[{Chuang et~al.(2020)Chuang, Robinson, Lin, Torralba, and Jegelka}]{debiased}
Chuang, C.; Robinson, J.; Lin, Y.; Torralba, A.; and Jegelka, S. 2020.
\newblock Debiased Contrastive Learning.
\newblock In Larochelle, H.; Ranzato, M.; Hadsell, R.; Balcan, M.; and Lin, H., eds., \emph{Advances in Neural Information Processing Systems 33: Annual Conference on Neural Information Processing Systems 2020, NeurIPS 2020, December 6-12, 2020, virtual}.

\bibitem[{Collins et~al.(2018)Collins, Tkaczyk, Aizawa, and Beel}]{position-bias}
Collins, A.; Tkaczyk, D.; Aizawa, A.; and Beel, J. 2018.
\newblock A Study of Position Bias in Digital Library Recommender Systems.
\newblock \emph{CoRR}, abs/1802.06565.

\bibitem[{Covington, Adams, and Sargin(2016)}]{youtube}
Covington, P.; Adams, J.; and Sargin, E. 2016.
\newblock Deep Neural Networks for YouTube Recommendations.
\newblock In \emph{Proceedings of the 10th ACM Conference on Recommender Systems}, RecSys '16, 191–198. New York, NY, USA: Association for Computing Machinery.
\newblock ISBN 9781450340359.

\bibitem[{Craswell et~al.(2008)Craswell, Zoeter, Taylor, and Ramsey}]{aecc}
Craswell, N.; Zoeter, O.; Taylor, M.; and Ramsey, B. 2008.
\newblock An Experimental Comparison of Click Position-Bias Models.
\newblock In \emph{Proceedings of the 2008 International Conference on Web Search and Data Mining}, WSDM '08, 87–94. New York, NY, USA: Association for Computing Machinery.
\newblock ISBN 9781595939272.

\bibitem[{Dacrema, Cremonesi, and Jannach(2019)}]{RecSys19Evaluation}
Dacrema, M.~F.; Cremonesi, P.; and Jannach, D. 2019.
\newblock Are We Really Making Much Progress? A Worrying Analysis of Recent Neural Recommendation Approaches.
\newblock In \emph{Proceedings of the 13th ACM Conference on Recommender Systems}, RecSys '19.

\bibitem[{Ding et~al.(2020)Ding, Quan, Yao, Li, and Jin}]{robustsample}
Ding, J.; Quan, Y.; Yao, Q.; Li, Y.; and Jin, D. 2020.
\newblock Simplify and Robustify Negative Sampling for Implicit Collaborative Filtering.
\newblock In Larochelle, H.; Ranzato, M.; Hadsell, R.; Balcan, M.; and Lin, H., eds., \emph{Advances in Neural Information Processing Systems}, volume~33, 1094--1105. Curran Associates, Inc.

\bibitem[{Gutmann and Hyvärinen(2010)}]{nce}
Gutmann, M.; and Hyvärinen, A. 2010.
\newblock Noise-contrastive estimation: A new estimation principle for unnormalized statistical models.
\newblock In Teh, Y.~W.; and Titterington, M., eds., \emph{Proceedings of the Thirteenth International Conference on Artificial Intelligence and Statistics}, volume~9 of \emph{Proceedings of Machine Learning Research}, 297--304. Chia Laguna Resort, Sardinia, Italy: PMLR.

\bibitem[{He et~al.(2020)He, Deng, Wang, Li, Zhang, and Wang}]{lightgcn}
He, X.; Deng, K.; Wang, X.; Li, Y.; Zhang, Y.; and Wang, M. 2020.
\newblock LightGCN: Simplifying and Powering Graph Convolution Network for Recommendation.
\newblock In \emph{Proceedings of the 43rd International ACM SIGIR Conference on Research and Development in Information Retrieval}, SIGIR '20, 639–648. New York, NY, USA: Association for Computing Machinery.
\newblock ISBN 9781450380164.

\bibitem[{He et~al.(2017)He, Liao, Zhang, Nie, Hu, and Chua}]{ncf}
He, X.; Liao, L.; Zhang, H.; Nie, L.; Hu, X.; and Chua, T. 2017.
\newblock Neural Collaborative Filtering.
\newblock In Barrett, R.; Cummings, R.; Agichtein, E.; and Gabrilovich, E., eds., \emph{Proceedings of the 26th International Conference on World Wide Web, {WWW} 2017, Perth, Australia, April 3-7, 2017}, 173--182. {ACM}.

\bibitem[{He et~al.(2016)He, Zhang, Kan, and Chua}]{eals}
He, X.; Zhang, H.; Kan, M.-Y.; and Chua, T.-S. 2016.
\newblock Fast Matrix Factorization for Online Recommendation with Implicit Feedback.
\newblock In \emph{Proceedings of the 39th International ACM SIGIR Conference on Research and Development in Information Retrieval}, SIGIR '16, 549–558. New York, NY, USA: Association for Computing Machinery.
\newblock ISBN 9781450340694.

\bibitem[{Hsieh et~al.(2017)Hsieh, Yang, Cui, Lin, Belongie, and Estrin}]{cml}
Hsieh, C.-K.; Yang, L.; Cui, Y.; Lin, T.-Y.; Belongie, S.; and Estrin, D. 2017.
\newblock Collaborative Metric Learning.
\newblock In \emph{Proceedings of the 26th International Conference on World Wide Web}, WWW '17, 193–201. Republic and Canton of Geneva, CHE: International World Wide Web Conferences Steering Committee.
\newblock ISBN 9781450349130.

\bibitem[{Hu, Koren, and Volinsky(2008{\natexlab{a}})}]{hu2008collaborative}
Hu, Y.; Koren, Y.; and Volinsky, C. 2008{\natexlab{a}}.
\newblock Collaborative filtering for implicit feedback datasets.
\newblock In \emph{ICDM'08}.

\bibitem[{Hu, Koren, and Volinsky(2008{\natexlab{b}})}]{ials@hu2008}
Hu, Y.; Koren, Y.; and Volinsky, C. 2008{\natexlab{b}}.
\newblock Collaborative Filtering for Implicit Feedback Datasets.
\newblock In \emph{2008 Eighth IEEE International Conference on Data Mining}, 263--272.

\bibitem[{Jean et~al.(2015)Jean, Cho, Memisevic, and Bengio}]{sample-softmax}
Jean, S.; Cho, K.; Memisevic, R.; and Bengio, Y. 2015.
\newblock On Using Very Large Target Vocabulary for Neural Machine Translation.
\newblock In \emph{Proceedings of the 53rd Annual Meeting of the Association for Computational Linguistics and the 7th International Joint Conference on Natural Language Processing of the Asian Federation of Natural Language Processing, {ACL} 2015, July 26-31, 2015, Beijing, China, Volume 1: Long Papers}, 1--10. The Association for Computer Linguistics.

\bibitem[{Jin et~al.(2021{\natexlab{a}})Jin, Li, Gao, Liu, Chen, and Zhou}]{jin@linear}
Jin, R.; Li, D.; Gao, J.; Liu, Z.; Chen, L.; and Zhou, Y. 2021{\natexlab{a}}.
\newblock Towards a Better Understanding of Linear Models for Recommendation.
\newblock In \emph{Proceedings of the 27th ACM SIGKDD Conference on Knowledge Discovery \&amp; Data Mining}, KDD '21, 776–785. New York, NY, USA: Association for Computing Machinery.
\newblock ISBN 9781450383325.

\bibitem[{Jin et~al.(2021{\natexlab{b}})Jin, Li, Mudrak, Gao, and Liu}]{Jin@AAAI21}
Jin, R.; Li, D.; Mudrak, B.; Gao, J.; and Liu, Z. 2021{\natexlab{b}}.
\newblock On Estimating Recommendation Evaluation Metrics under Sampling.
\newblock \emph{Proceedings of the AAAI Conference on Artificial Intelligence}, 4147--4154.

\bibitem[{Krichene and Rendle(2020)}]{walid@sample}
Krichene, W.; and Rendle, S. 2020.
\newblock On Sampled Metrics for Item Recommendation.
\newblock In \emph{Proceedings of the 26th ACM SIGKDD International Conference on Knowledge Discovery \&amp; Data Mining}, KDD '20, 1748–1757. New York, NY, USA: Association for Computing Machinery.
\newblock ISBN 9781450379984.

\bibitem[{Li, Bhargavi, and Ravipati(2023)}]{li2023impressioninformed}
Li, D.; Bhargavi, D.; and Ravipati, V.~S. 2023.
\newblock Impression-Informed Multi-Behavior Recommender System: A Hierarchical Graph Attention Approach.
\newblock arXiv:2309.03169.

\bibitem[{Li et~al.(2020)Li, Jin, Gao, and Liu}]{Li@KDD20}
Li, D.; Jin, R.; Gao, J.; and Liu, Z. 2020.
\newblock On Sampling Top-K Recommendation Evaluation.
\newblock In \emph{Proceedings of the 26th ACM SIGKDD International Conference on Knowledge Discovery \& Data Mining}, KDD '20.

\bibitem[{Li et~al.(2023{\natexlab{a}})Li, Jin, Liu, Ren, Gao, and Liu}]{li@acm23}
Li, D.; Jin, R.; Liu, Z.; Ren, B.; Gao, J.; and Liu, Z. 2023{\natexlab{a}}.
\newblock On Item-Sampling Evaluation for Recommender System.
\newblock \emph{ACM Trans. Recomm. Syst.}
\newblock Just Accepted.

\bibitem[{Li et~al.(2023{\natexlab{b}})Li, Jin, Liu, Ren, Gao, and Liu}]{dong@2023aaai}
Li, D.; Jin, R.; Liu, Z.; Ren, B.; Gao, J.; and Liu, Z. 2023{\natexlab{b}}.
\newblock Towards Reliable Item Sampling for Recommendation Evaluation.
\newblock In \emph{Thirty-Seventh {AAAI} Conference on Artificial Intelligence, {AAAI} 2023}. {AAAI} Press.

\bibitem[{Li et~al.(2022)Li, Shen, Jin, Mao, Wang, and Chen}]{li2022generationaugmented}
Li, D.; Shen, Y.; Jin, R.; Mao, Y.; Wang, K.; and Chen, W. 2022.
\newblock Generation-Augmented Query Expansion For Code Retrieval.
\newblock arXiv:2212.10692.

\bibitem[{Lian, Liu, and Chen(2020)}]{importance}
Lian, D.; Liu, Q.; and Chen, E. 2020.
\newblock Personalized Ranking with Importance Sampling.
\newblock In \emph{Proceedings of The Web Conference 2020}, WWW '20, 1093–1103. New York, NY, USA: Association for Computing Machinery.
\newblock ISBN 9781450370233.

\bibitem[{Liang et~al.(2016)Liang, Charlin, McInerney, and Blei}]{user-bias}
Liang, D.; Charlin, L.; McInerney, J.; and Blei, D.~M. 2016.
\newblock Modeling User Exposure in Recommendation.
\newblock In \emph{Proceedings of the 25th International Conference on World Wide Web}, WWW '16, 951–961. Republic and Canton of Geneva, CHE: International World Wide Web Conferences Steering Committee.
\newblock ISBN 9781450341431.

\bibitem[{Liang et~al.(2018)Liang, Krishnan, Hoffman, and Jebara}]{multi-vae}
Liang, D.; Krishnan, R.~G.; Hoffman, M.~D.; and Jebara, T. 2018.
\newblock Variational Autoencoders for Collaborative Filtering.
\newblock In Champin, P.; Gandon, F.; Lalmas, M.; and Ipeirotis, P.~G., eds., \emph{Proceedings of the 2018 World Wide Web Conference on World Wide Web, {WWW} 2018, Lyon, France, April 23-27, 2018}, 689--698. {ACM}.

\bibitem[{Liu et~al.(2020)Liu, Cheng, Dong, He, Pan, and Ming}]{expose-bias}
Liu, D.; Cheng, P.; Dong, Z.; He, X.; Pan, W.; and Ming, Z. 2020.
\newblock A General Knowledge Distillation Framework for Counterfactual Recommendation via Uniform Data.
\newblock In \emph{Proceedings of the 43rd International ACM SIGIR Conference on Research and Development in Information Retrieval}, SIGIR '20, 831–840. New York, NY, USA: Association for Computing Machinery.
\newblock ISBN 9781450380164.

\bibitem[{Mao et~al.(2021)Mao, Zhu, Wang, Dai, Dong, Xiao, and He}]{simplex}
Mao, K.; Zhu, J.; Wang, J.; Dai, Q.; Dong, Z.; Xiao, X.; and He, X. 2021.
\newblock SimpleX: A Simple and Strong Baseline for Collaborative Filtering.
\newblock In \emph{Proceedings of the 30th ACM International Conference on Information \&amp; Knowledge Management}, CIKM '21, 1243–1252. New York, NY, USA: Association for Computing Machinery.
\newblock ISBN 9781450384469.

\bibitem[{Marlin et~al.(2007)Marlin, Zemel, Roweis, and Slaney}]{select-bias}
Marlin, B.~M.; Zemel, R.~S.; Roweis, S.~T.; and Slaney, M. 2007.
\newblock Collaborative Filtering and the Missing at Random Assumption.
\newblock In Parr, R.; and van~der Gaag, L.~C., eds., \emph{{UAI} 2007, Proceedings of the Twenty-Third Conference on Uncertainty in Artificial Intelligence, Vancouver, BC, Canada, July 19-22, 2007}, 267--275. {AUAI} Press.

\bibitem[{Ning and Karypis(2011)}]{slim}
Ning, X.; and Karypis, G. 2011.
\newblock SLIM: Sparse Linear Methods for Top-N Recommender Systems.
\newblock In \emph{2011 IEEE 11th International Conference on Data Mining}, 497--506.

\bibitem[{Oord, Li, and Vinyals(2018)}]{infonce}
Oord, A. v.~d.; Li, Y.; and Vinyals, O. 2018.
\newblock Representation learning with contrastive predictive coding.
\newblock \emph{arXiv preprint arXiv:1807.03748}.

\bibitem[{Rendle(2021)}]{rendle2021item}
Rendle, S. 2021.
\newblock Item recommendation from implicit feedback.
\newblock In \emph{Recommender Systems Handbook}, 143--171. Springer.

\bibitem[{Rendle and Freudenthaler(2014)}]{aobpr}
Rendle, S.; and Freudenthaler, C. 2014.
\newblock Improving Pairwise Learning for Item Recommendation from Implicit Feedback.
\newblock In \emph{Proceedings of the 7th ACM International Conference on Web Search and Data Mining}, WSDM '14, 273–282. New York, NY, USA: Association for Computing Machinery.
\newblock ISBN 9781450323512.

\bibitem[{Rendle et~al.(2009)Rendle, Freudenthaler, Gantner, and Schmidt-Thieme}]{bpr}
Rendle, S.; Freudenthaler, C.; Gantner, Z.; and Schmidt-Thieme, L. 2009.
\newblock BPR: Bayesian Personalized Ranking from Implicit Feedback.
\newblock In \emph{Proceedings of the Twenty-Fifth Conference on Uncertainty in Artificial Intelligence}, UAI '09, 452–461. AUAI Press.
\newblock ISBN 9780974903958.

\bibitem[{Rendle et~al.(2020)Rendle, Krichene, Zhang, and Anderson}]{ncfmf}
Rendle, S.; Krichene, W.; Zhang, L.; and Anderson, J. 2020.
\newblock Neural Collaborative Filtering vs. Matrix Factorization Revisited.
\newblock In \emph{Proceedings of the 14th ACM Conference on Recommender Systems}, RecSys '20, 240–248. New York, NY, USA: Association for Computing Machinery.
\newblock ISBN 9781450375832.

\bibitem[{Rendle et~al.(2021)Rendle, Krichene, Zhang, and Koren}]{ials++}
Rendle, S.; Krichene, W.; Zhang, L.; and Koren, Y. 2021.
\newblock iALS++: Speeding up Matrix Factorization with Subspace Optimization.

\bibitem[{Rendle et~al.(2022)Rendle, Krichene, Zhang, and Koren}]{ials_revisiting}
Rendle, S.; Krichene, W.; Zhang, L.; and Koren, Y. 2022.
\newblock Revisiting the Performance of IALS on Item Recommendation Benchmarks.
\newblock In \emph{Proceedings of the 16th ACM Conference on Recommender Systems}, RecSys '22, 427–435. New York, NY, USA: Association for Computing Machinery.
\newblock ISBN 9781450392785.

\bibitem[{Rendle, Zhang, and Koren(2019)}]{difficulty@rendle}
Rendle, S.; Zhang, L.; and Koren, Y. 2019.
\newblock On the Difficulty of Evaluating Baselines: {A} Study on Recommender Systems.
\newblock \emph{CoRR}, abs/1905.01395.

\bibitem[{Saito(2020)}]{unbiased-pair}
Saito, Y. 2020.
\newblock Unbiased Pairwise Learning from Biased Implicit Feedback.
\newblock In \emph{Proceedings of the 2020 ACM SIGIR on International Conference on Theory of Information Retrieval}, ICTIR '20, 5–12. New York, NY, USA: Association for Computing Machinery.
\newblock ISBN 9781450380676.

\bibitem[{Saito et~al.(2020)Saito, Yaginuma, Nishino, Sakata, and Nakata}]{unbiased-rec-miss}
Saito, Y.; Yaginuma, S.; Nishino, Y.; Sakata, H.; and Nakata, K. 2020.
\newblock Unbiased Recommender Learning from Missing-Not-At-Random Implicit Feedback.
\newblock In \emph{Proceedings of the 13th International Conference on Web Search and Data Mining}, WSDM '20, 501–509. New York, NY, USA: Association for Computing Machinery.
\newblock ISBN 9781450368223.

\bibitem[{Steck(2010)}]{steck-debias}
Steck, H. 2010.
\newblock Training and Testing of Recommender Systems on Data Missing Not at Random.
\newblock In \emph{Proceedings of the 16th ACM SIGKDD International Conference on Knowledge Discovery and Data Mining}, KDD '10, 713–722. New York, NY, USA: Association for Computing Machinery.
\newblock ISBN 9781450300551.

\bibitem[{Steck(2019)}]{ease}
Steck, H. 2019.
\newblock Embarrassingly Shallow Autoencoders for Sparse Data.
\newblock \emph{CoRR}, abs/1905.03375.

\bibitem[{Steck(2020)}]{edlae}
Steck, H. 2020.
\newblock Autoencoders that don't overfit towards the Identity.
\newblock In Larochelle, H.; Ranzato, M.; Hadsell, R.; Balcan, M.; and Lin, H., eds., \emph{Advances in Neural Information Processing Systems 33: Annual Conference on Neural Information Processing Systems 2020, NeurIPS 2020, December 6-12, 2020, virtual}.

\bibitem[{Tian, Krishnan, and Isola(2020)}]{multi-cl}
Tian, Y.; Krishnan, D.; and Isola, P. 2020.
\newblock Contrastive Multiview Coding.
\newblock In Vedaldi, A.; Bischof, H.; Brox, T.; and Frahm, J., eds., \emph{Computer Vision - {ECCV} 2020 - 16th European Conference, Glasgow, UK, August 23-28, 2020, Proceedings, Part {XI}}, volume 12356 of \emph{Lecture Notes in Computer Science}, 776--794. Springer.

\bibitem[{Veli{\v{c}}kovi{\'c} et~al.(2018)Veli{\v{c}}kovi{\'c}, Cucurull, Casanova, Romero, Lio, and Bengio}]{gat}
Veli{\v{c}}kovi{\'c}, P.; Cucurull, G.; Casanova, A.; Romero, A.; Lio, P.; and Bengio, Y. 2018.
\newblock Graph attention networks.
\newblock In \emph{International Conference on Learning Representations}.

\bibitem[{Wang et~al.(2022)Wang, Yu, Ma, Zhang, Chen, Liu, and Ma}]{alignment-uniformity}
Wang, C.; Yu, Y.; Ma, W.; Zhang, M.; Chen, C.; Liu, Y.; and Ma, S. 2022.
\newblock Towards Representation Alignment and Uniformity in Collaborative Filtering.
\newblock In \emph{Proceedings of the 28th ACM SIGKDD Conference on Knowledge Discovery and Data Mining}, KDD '22, 1816–1825. New York, NY, USA: Association for Computing Machinery.
\newblock ISBN 9781450393850.

\bibitem[{Wang et~al.(2019)Wang, He, Wang, Feng, and Chua}]{ngcf}
Wang, X.; He, X.; Wang, M.; Feng, F.; and Chua, T. 2019.
\newblock Neural Graph Collaborative Filtering.
\newblock In Piwowarski, B.; Chevalier, M.; Gaussier, {\'{E}}.; Maarek, Y.; Nie, J.; and Scholer, F., eds., \emph{Proceedings of the 42nd International {ACM} {SIGIR} Conference on Research and Development in Information Retrieval, {SIGIR} 2019, Paris, France, July 21-25, 2019}, 165--174. {ACM}.

\bibitem[{Wu et~al.(2018)Wu, Lee, Li, Pan, and Zhang}]{cellpad}
Wu, J.; Lee, P. P.~C.; Li, Q.; Pan, L.; and Zhang, J. 2018.
\newblock CellPAD: Detecting Performance Anomalies in Cellular Networks via Regression Analysis.
\newblock In \emph{2018 IFIP Networking Conference (IFIP Networking) and Workshops}, 1--9.

\bibitem[{Wu et~al.(2021)Wu, Wang, Feng, He, Chen, Lian, and Xie}]{sgl-ed}
Wu, J.; Wang, X.; Feng, F.; He, X.; Chen, L.; Lian, J.; and Xie, X. 2021.
\newblock Self-supervised Graph Learning for Recommendation.
\newblock In Diaz, F.; Shah, C.; Suel, T.; Castells, P.; Jones, R.; and Sakai, T., eds., \emph{{SIGIR} '21: The 44th International {ACM} {SIGIR} Conference on Research and Development in Information Retrieval, Virtual Event, Canada, July 11-15, 2021}, 726--735. {ACM}.

\bibitem[{Wu, Ye, and Man(2023)}]{bot}
Wu, J.; Ye, X.; and Man, Y. 2023.
\newblock BotTriNet: A Unified and Efficient Embedding for Social Bots Detection via Metric Learning.
\newblock In \emph{2023 11th International Symposium on Digital Forensics and Security (ISDFS)}, 1--6.

\bibitem[{Yeh et~al.(2022)Yeh, Hong, Hsu, Liu, Chen, and LeCun}]{decoupling@eccv}
Yeh, C.; Hong, C.; Hsu, Y.; Liu, T.; Chen, Y.; and LeCun, Y. 2022.
\newblock Decoupled Contrastive Learning.
\newblock In Avidan, S.; Brostow, G.~J.; Ciss{\'{e}}, M.; Farinella, G.~M.; and Hassner, T., eds., \emph{Computer Vision - {ECCV} 2022 - 17th European Conference, Tel Aviv, Israel, October 23-27, 2022, Proceedings, Part {XXVI}}, volume 13686 of \emph{Lecture Notes in Computer Science}, 668--684. Springer.

\bibitem[{Ying et~al.(2018)Ying, He, Chen, Eksombatchai, Hamilton, and Leskovec}]{pinsage}
Ying, R.; He, R.; Chen, K.; Eksombatchai, P.; Hamilton, W.~L.; and Leskovec, J. 2018.
\newblock Graph Convolutional Neural Networks for Web-Scale Recommender Systems.
\newblock In Guo, Y.; and Farooq, F., eds., \emph{Proceedings of the 24th {ACM} {SIGKDD} International Conference on Knowledge Discovery {\&} Data Mining, {KDD} 2018, London, UK, August 19-23, 2018}, 974--983. {ACM}.

\bibitem[{Zhang et~al.(2019)Zhang, Yao, Sun, and Tay}]{zhang2019deep}
Zhang, S.; Yao, L.; Sun, A.; and Tay, Y. 2019.
\newblock Deep Learning Based Recommender System: A Survey and New Perspectives.
\newblock \emph{ACM Comput. Surv.}, 52(1).

\bibitem[{Zhang et~al.(2013)Zhang, Chen, Wang, and Yu}]{dynamic}
Zhang, W.; Chen, T.; Wang, J.; and Yu, Y. 2013.
\newblock Optimizing Top-n Collaborative Filtering via Dynamic Negative Item Sampling.
\newblock In \emph{Proceedings of the 36th International ACM SIGIR Conference on Research and Development in Information Retrieval}, SIGIR '13, 785–788. New York, NY, USA: Association for Computing Machinery.
\newblock ISBN 9781450320344.

\bibitem[{Zhou et~al.(2021)Zhou, Ma, Zhang, Zhou, and Yang}]{cl-rec}
Zhou, C.; Ma, J.; Zhang, J.; Zhou, J.; and Yang, H. 2021.
\newblock Contrastive Learning for Debiased Candidate Generation in Large-Scale Recommender Systems.
\newblock In \emph{Proceedings of the 27th ACM SIGKDD Conference on Knowledge Discovery \&amp; Data Mining}, KDD '21, 3985–3995. New York, NY, USA: Association for Computing Machinery.
\newblock ISBN 9781450383325.

\bibitem[{Zhu et~al.(2020)Zhu, He, Zhang, and Caverlee}]{unbiased-propensity}
Zhu, Z.; He, Y.; Zhang, Y.; and Caverlee, J. 2020.
\newblock Unbiased Implicit Recommendation and Propensity Estimation via Combinational Joint Learning.
\newblock RecSys '20, 551–556. New York, NY, USA: Association for Computing Machinery.
\newblock ISBN 9781450375832.

\end{thebibliography}

\appendix
\clearpage
\section{Related Work}
\label{sec:related}
\subsection{Objectives of Implicit Collaborative Filtering}

Implicit feedback has been popular for decades in Recommender System \citep{rendle2021item} since Hu et al. first proposed iALS \citep{ials@hu2008}, where a second-order pointwise objective - Mean Square Error (MSE) was adopted to optimize user and item embeddings alternatively. Due to its effectiveness and efficiency, there is a large number of works that take the MSE or its variants as their objectives, spanning from Matrix Factorization (MF) based models \citep{ials_revisiting,ials++,eals}, to regression-based models, including SLIM \citep{slim}, EASE \citep{ease,edlae}, etc. He et al. treat collaborative filtering  as a binary classification task and apply the pointwise objective - Binary Cross-Entropy Loss (BCE) onto it \citep{ncf}. CML utilized the pairwise hinge loss onto the collaborative filtering scenario \citep{cml}. MultiVAE \citep{multi-vae} utilizes multinomial likelihood estimation. Rendle et al. proposed a Bayesian perspective pairwise ranking loss - BPR, in the seminal work \citep{bpr}. YouTubeNet posed a recommendation as an extreme multiclass classification problem and apply the Softmax cross-entropy (Softmax) \citep{youtube,sample-softmax}. Recently, inspired by the largely used contrastive loss in the computer vision area, \citep{simplex} proposed a first-order based Cosine Contrastive Loss (CCL), where they maximize the cosine similarity between the positive pairs (between users and items) and minimize the similarity below some manually selected margin.

\subsection{Contrastive Learning}
Recently, Contrastive Learning (CL) \citep{SimCLR,infonce} has become a prominent optimizing framework in deep learning and machine learning \cite{li2022generationaugmented,cellpad,bot,li2023impressioninformed}. The motivation behind CL is to learn representations by contrasting positive and negative pairs as well as maximize the positive pairs, including data augmentation \citep{SimCLR} and multi-view representations \citep{multi-cl}, etc. 
Chuang et al. proposed a new unsupervised contrastive representation learning framework, targeting to minimize the bias introduced by the selection of the negative samples. This debiased objective consistently improved its counterpart - the biased one in various benchmarks \citep{debiased}. Lately, by introducing the margin between positive and negative samples, Belghazi et al. present a supervised contrastive learning framework that is robust to biases \citep{unbiased}.
In addition, a mutual information estimator - MINE \citep{MINE}, closely related to the contrastive learning loss InfoNCE\citep{infonce}, has demonstrated its efficiency in various optimizing settings. \citep{alignment-uniformity} study the alignment and uniformity of user, item embeddings from the perspective of contrastive learning in the recommendation. CLRec \citep{cl-rec} design a new contrastive learning loss which is equivalent to using inverse propensity weighting to reduce exposure bias of a large-scale system. 



\subsection{Negative Sampling Strategies for Item Recommendation}
In most real-world scenarios, only positive feedback is available which brings demand for negative signals to avoid trivial solutions during training recommendation models. Apart from some non-sampling frameworks like iALS \citep{ials@hu2008}, Multi-VAE \citep{multi-vae}, the majority of models \citep{negative_sampling_review,lightgcn,ncf,bpr,rendle2021item,sampling_strategy} would choose to sample negative items for efficiency consideration. Uniform sampling is the most popular one \citep{bpr} which assumes uniform prior distribution. Importance sampling is another popular choice \citep{importance}, which chooses negative according to their frequencies. Adaptive sampling strategy keep tracking and picking the negative samples with higher scores \citep{aobpr,dynamic}. Another similar strategy is SRNS \citep{robustsample}, which tries to find and optimize false negative samples 
In addition, NCE \citep{nce} approach for negative sampling has also been adopted in the recommendation system\citep{sampling_strategy}. 


\subsection{Bias and Debias in Recommendation}
In the recommender system, there are various different bias \citep{bias-debias}, 
including item exposure bias, popularity bias, position bias, selection bias, etc.
These issues have to be carefully handled otherwise it would lead to unexpected results, such as inconsistency of online and offline performance, etc. Marlin et al. \citep{select-bias} validate the existence of selection bias by conducting a user survey, which indicates the discrepancy between the observed rating data and all data. Since users are generally exposed to a small set of items, unobserved data doesn't mean negative preference, this is so-called exposure bias \citep{expose-bias}. Position bias \citep{position-bias} is another commonly 
encountered one, where users are more likely to select items that display in a desirable place or higher position on a list.

Steck et al. \citep{steck-debias} propose an unbiased metric to deal with selection bias. iALS \citep{ials@hu2008} put some weights on unobserved data which helps improve the performance and deal with exposure bias. In addition, the sampling strategy would naturally diminish the exposure bias during model training \citep{bias-debias,bpr}. EXMF \citep{user-bias} is a user exposure simulation model that can also help reduce exposure bias.
To reduce position bias, some models \citep{dbcl,aecc} make assumptions about user behaviors and estimate true relevance instead of directly using clicked data. \citep{unbiased-pair} proposed an unbiased pairwise BPR estimator and further provide a practical technique to reduce variance. Another widely used method to correct bias is the utilization of propensity score \citep{unbiased-propensity,unbiased-rec-miss} where each item is equiped with some position weight.

\section{reproducibility}\label{app:reproduce}

\begin{table}
\caption{Statistics of the datasets.}
\label{tab:dataset}
\begin{tabular}{c|c|c|c}
\hline { Dataset } & User \# & Item \# & Interaction \#  \\
\hline 
\hline Yelp2018 & 31,668 & 38,048 & $1,561,406$  \\
\hline Gowalla & 29,858 & 40,981 & $1,027,370$  \\
\hline Amazon-Books & 52,643 & 91,599 & $2,984,108$  \\
\hline
\end{tabular}
\end{table}

\subsection{Implement details}


In \cref{tab:mine-repro}, we list the details of the hyperparameters for reproducing the results of MINE+ (\cref{eq:mine+}) in \cref{tab:main,tab:debias}.

In \cref{tab:ccl-repro}, we list the details of the hyperparameters for reproducing the results of debiased $CCL$ in \cref{tab:debias}.

\begin{table}[]
\caption{Hyperparameter details of $MINE+$ loss}
\label{tab:mine-repro}
\small{
\begin{tabular}{|c|c|c|c|}
\hline
\multicolumn{1}{|l|}{} & Yelp2018 & Gowalla & Amazon-Books \\ \hline
negative weight        & 1.1      & 1.2     & 1.1          \\ \hline
temperature            & 0.5      & 0.4     & 0.4          \\ \hline
regularization         & 1       & 1     & 0.01         \\ \hline
number of negative     & 800      & 800     & 800          \\ \hline
\end{tabular}
}
\end{table}

\begin{table}[]
\caption{Hyperparameter details of debiased $CCL$ loss}
\label{tab:ccl-repro}
\small{
\begin{tabular}{|c|c|c|c|}
\hline
\multicolumn{1}{|l|}{} & Yelp2018 & Gowalla & Amazon-Books \\ \hline
negative weight        & 0.4      & 0.7     & 0.6          \\ \hline
margin                 & 0.9      & 0.9     & 0.4          \\ \hline
regularization         & -9       & -9      & -9           \\ \hline
number of negative     & 800      & 800     & 800          \\ \hline
number of positive     & 10       & 20      & 50           \\ \hline
\end{tabular}
}
\end{table}

\section{MASE, iALS, EASE and their Debiasness}

Let us denote the generic single pointwise loss function $l^+(\hat{y}_{ui})$ and $l^-(\hat{y}_{ui})$, which measure how close the individual positive (negative) item score to their ideal target value. 

\noindent{\bf MSE single pointwise loss:}
Mean-Squared-Error (MSE) is one of the most widely used recommendation loss functions.  Indeed, all the earlier collaborative filtering (Matrix factorization) models and pointwise losses were  based on MSE (with different regularizations). Some recent linear models, such as SLIM~\cite{slim} and EASE~\cite{ease}, are also based on MSE. 
Its single pointwise loss function is denoted as: 
\begin{equation}
\begin{cases}
    & l^+_{mse}(\hat{y}_{ui})=(1-\hat{y}_{ui})^2 \\
    & l^-_{mse}(\hat{y}_{ui})=(\hat{y}_{ui})^2
\end{cases}
\end{equation}

\bdefin {\bf (Ideal Pointwise Loss)}
The ideal pointwise loss function for recommendation models is the expected single pointwise loss for all positive items (sampled from $\rho_u(+)=
\tau^+$) and for negative items (sampled from $\rho_u(-)=
\tau^-$):  
\begin{small}
\begin{equation}
    \begin{split}
    \overline{\mathcal{L}}_{point} = \E_{u} \Bigg(\tau_u^+\E_{i\sim p_u^+} l^+(\hat{y}_{ui}) +
    \lambda\tau_u^-\E_{j\sim p_u^-} l^-(\hat{y}_{uj}) \Bigg)
    \end{split}
\end{equation}
\end{small}
\edefin

Why this is ideal? When the optimization reaches the potential minimal, we have all positive items $i$ and all the negative items $j$ reaching all the minimum of individual loss ($l^+$ and $l^-$).  
Here, $\lambda$ help adjusts the balance between the positive and negative losses, similar to iALS ~\cite{hu2008collaborative,ials_revisiting} and CCL~\cite{simplex}. 
However, since $p_u^+$ and $p_u^-$ is unknown, we utilize the same debiasing approach as InfoNCE ~\cite{infonce} for the debiased MSE: 

\bdefin {\bf (Debiased Ideal pointwise loss)}
\begin{small}
\begin{equation}
    \begin{split}
    &\mathcal{L}^{Debiased}_{point}= \E_{u} \Bigl( \tau_u^+\E_{i\sim p_u^+} l^+(\hat{y}_{ui}) + \tau_u^-\lambda \cdot \E_{j\sim p_u^-} l^-(\hat{y}_{ui}) \Bigr)\\
    & = \E_{u} \Big(\tau_u^+\E_{i\sim p_u^+}l^+(\hat{y}_{ui}) + \lambda\Big( 
    \E_{j\sim p_u}l^-(\hat{y}_{ui})  - \tau_u^+ \E_{k\sim k^+_u}l^-(\hat{y}_{uj}) \Big)\Big)
    \end{split}
\end{equation}
\end{small}
\edefin

In practice, assuming for each positive item, we sample $N$ positive items and $M$ negative items, then the empirical loss can be written as: 

\begin{small}
\begin{equation}
    \begin{split}
&\widetilde{\mathcal{L}}^{Debiased}_{point}= \E_{u} E_{i \sim p_u^+} \Bigl( \tau_u^+l^+(\hat{y}_{ui}) + \\ 
    &\lambda \Bigl( \frac{1}{N} \sum_{j=1;j\sim p_u}^N l^-(\hat{y}_{uj})- \tau_u^+\frac{1}{M} \sum_{k=1;k\sim p_u^+}^M l^-(\hat{y}_{uk}) \Bigr) \Bigr)
    \end{split}
\end{equation} 
\end{small}



Here, we investigate how the debiased MSE loss will impact the solution of two (arguably) most popular linear recommendation models, iALS ~\cite{hu2008collaborative,ials_revisiting} and EASE~\cite{ease}. 
Rather surprisingly, we found the solvers of both models can absorb the debiased loss under their existing framework with reasonable conditions (details below). In other words, both iALS and EASE can be considered to be inherently debiased. 

To obtain the aforementioned insights, we first transform the debiased loss into the summation form being used in iALS and EASE. 
\begin{scriptsize}
\begin{equation*}
\begin{split}
&\mathcal{L}^{Debiased}_{mse} \approx 
\E_{u} \Bigg[ \frac{\tau^+_u} {|\mathcal{I}^+_u|}\sum\limits_{i\in \mathcal{I}_u^+}(\hat{y}_{ui}-1)^2 + \lambda\Bigg(\frac{1}{|\mathcal{I}|} \sum\limits_{t\in\mathcal{I}} \hat{y}_{ut}^2 -\frac{\tau^+_u}{|\mathcal{I}_u^+|}\sum\limits_{q\in \mathcal{I}_u^+}\hat{y}^2_{uq} \Bigg)\Bigg]\\
&=\sum\limits_{u} \Bigg[ \frac{1}{|\mathcal{I}|} c_u \sum\limits_{i\in \mathcal{I}_u^+}(\hat{y}_{ui}-1)^2 \  \ \ \ \ \ \ \ \ \ \ \ \ \ \text{where, }  c_u=\frac{|\mathcal{I}|}{|\mathcal{I}_u^+|} \tau_u^+
\\
&+ \lambda\Bigg(\frac{1}{|\mathcal{I}|} \sum\limits_{t\in\mathcal{I}} \hat{y}_{ut}^2 -\frac{1}{|\mathcal{I}|} c_u \sum\limits_{q\in \mathcal{I}_u^+}\hat{y}^2_{uq} \Bigg)\Bigg]\\
&\propto \sum\limits_{u} \Bigg[ c_u \sum\limits_{i\in \mathcal{I}_u^+}(\hat{y}_{ui}-1)^2 + \lambda\Bigg(\sum\limits_{t\in\mathcal{I}} \hat{y}_{ut}^2 - c_u\sum\limits_{q\in \mathcal{I}_u^+}\hat{y}^2_{uq} \Bigg)\Bigg]\\
&=\sum\limits_{u} \Bigg[ \sum\limits_{i\in \mathcal{I}_u^+}[c_u(\hat{y}_{ui}-1)^2 -c_u\lambda \hat{y}^2_{uq}]+ \lambda\sum\limits_{t\in\mathcal{I}} \hat{y}_{ut}^2 \Bigg]
\end{split}
\end{equation*}
\end{scriptsize}

\noindent{\bf Debiased iALS:}
Following the original iALS paper \citep{ials@hu2008}, and the latest revised iALS \citep{ials_revisiting,ials++}, the objective of iALS is given by \citep{ials_revisiting}:

\begin{scriptsize}
\begin{equation}
    \begin{split}
    \mathcal{L}_{iALS}&=\sum\limits_{(u,i)\in S} (\hat{y}(u,i)-1)^2 + \alpha_0\sum\limits_{u\in U}\sum\limits_{i\in I}\hat{y}(u,i)^2\\
    &+\lambda\Bigg( \sum\limits_{u\in U}(|\mathcal{I}_u^+|+\alpha_0|\mathcal{I}|)^{\nu}||\mathbf{w}_u||^2 + \sum\limits_{i\in I}(|U_i^+|+\alpha_0|\mathcal{U}|)^{\nu}||\mathbf{h}_i||^2\Bigg)
    \end{split}
\end{equation}
\end{scriptsize}
where $\lambda$ is global regularization weight (hyperparameter) and $\alpha_0$ is unobserved weight (hyperparameter). $\mu$ is generally set to be $1$. 
Also, $\mathbf{w}_u$ and $\mathbf{h}_i$ are user and item vectors for user $u$ and item $i$, respectively. 

Now, we consider applying the debiased MSE loss to replace the original MSE loss (the first line in $\mathcal{L}_{iALS}$) with the second line unchanged: 

\begin{scriptsize}
\begin{equation*}
\begin{split}
&\mathcal{L}^{Debiased}_{iALS}
=\sum\limits_{u\in U} \Bigg[ \sum\limits_{i\in \mathcal{I}_u^+}[c_u(\hat{y}_{ui}-1)^2 -c_u\cdot \alpha_0 \cdot \hat{y}^2_{ui}]+ \alpha_0\sum\limits_{t\in I} \hat{y}_{ut}^2 \Bigg]\\
 &+\lambda\Bigg( \sum\limits_{u\in U}(|\mathcal{I}_u^+|+\alpha_0|\mathcal{I}|)^{\nu}||\mathbf{w}_u||^2 + \sum\limits_{i\in I}(|\mathcal{U}_i^+|+\alpha_0|\mathcal{U}|)^{\nu}||\mathbf{h}_i||^2\Bigg)
\end{split}
\end{equation*}
\end{scriptsize}

Then we have the following conclusion:
\bthm\label{th1}
For any debiased iALS loss $\mathcal{L}^{Debiased}_{iALS}$ with parameters $\alpha_0$ and $\lambda$ with constant $c_u$ for all users, there are original iALS loss with parameters  $\alpha_0^\prime$ and $\lambda^\prime$, which have the same closed form solutions (up to a constant factor) for fixing item vectors and user vectors, respectively.
\ethm 

\begin{proof}
To derive the alternating least square solution, we have: 
\begin{itemize}
    \item Fixing item vectors, optimizing: 
    \begin{scriptsize}
    \begin{equation*}
    \begin{split}
        \mathcal{L}_u &=||\sqrt{c_u}{(H^{u}_S)}^T\mathbf{w}_u - \sqrt{c_u}\mathbf{y}_S||^2 + ||\sqrt{\alpha_0}H^T\mathbf{w}_u||^2 + ||\sqrt{\lambda_u}I \cdot \mathbf{w}_u||^2\\
        &-||\sqrt{c_u\alpha_0} {(H^{u}_S)}^T\mathbf{w}_u||^2, \ \ \ \ where \ \ \ \ \lambda_u = \lambda(|\mathcal{I}_u^+|+\alpha_0|\mathcal{I}|)^{\nu} 
    \end{split}
\end{equation*}
\end{scriptsize}
Note that the observed item matrix for user $u$ as $H^u_S$:

\begin{scriptsize}
\begin{equation}
H^u_S=\begin{bmatrix}
\vrule & \vrule &        & \vrule\\
\mathbf{h}_1 & \mathbf{h}_s & \cdots & \mathbf{h}_{|\mathcal{I}_u^+|}\\
\vrule & \vrule &        & \vrule\\
\end{bmatrix}\in \mathbb{R}^{k\times |\mathcal{I}_u^+|},\quad s\in \mathcal{I}_u^+
\end{equation}
\end{scriptsize}

$H$ is the entire item matrix for all items. 
$y_S$ are all $1$ column (the same row dimension as $H^u_S$.

    \item Fixing User Vectors, optimizing: 
    \begin{scriptsize}
 \begin{equation*}
    \begin{split}
        \mathcal{L}_i &=||{(W^{i}_S \sqrt{C_u})}^T\mathbf{h}_i - \sqrt{C_u}\mathbf{y}_S||^2 + ||\sqrt{\alpha_0}W^T\mathbf{h}_i||^2 + ||\sqrt{\lambda_i}I \cdot \mathbf{h}_i||^2\\
        &-||\sqrt{\alpha_0} {(\sqrt{C_u}W^{i}_S)}^T\mathbf{h}_i||^2, \ \ \ \ where \ \ \lambda_i = \lambda(|\mathcal{U}_i^+|+\alpha_0|\mathcal{U}|)^{\nu}
    \end{split}
\end{equation*}
\end{scriptsize}
Here, $W^{i}_S$ are observed user matrix for item $i$, and $W$ are the entire user matrix, $C_u=diag(c_u)$ a $|\mathcal{U}|\times|\mathcal{U}|$ diagonal matrix  with $c_u$ on the diagonal.  
\end{itemize}

Solving $\mathcal{L}_u$ and $\mathcal{L}_i$, we have the following closed form solutions:
\begin{scriptsize}
\begin{equation*}
    \begin{split}
        & \mathbf{w}_u^*=\Big(c_u (1 -\alpha_0)H^{u}_S{(H^{u}_S)}^T + \alpha_0 HH^T + \lambda_u I\Big)^{-1}\cdot H^{u}_S\cdot \sqrt{c_u}\mathbf{y}_S \\ 
        & \mathbf{h}_i^*=\Big((1 -\alpha_0)W^{i}_SC_u{(W^{i}_S)}^T + \alpha_0 WW^T + \lambda_i I\Big)^{-1}\cdot W^{i}_S\cdot \sqrt{C_u}\mathbf{y}_S
    \end{split}
\end{equation*}
\end{scriptsize}

Assuming $c_u$ be a constant for all users, we get 

\begin{scriptsize}
\begin{equation*}
    \begin{split}
        & \mathbf{w}_u^* \varpropto \Big(H^{u}_S{(H^{u}_S)}^T + \frac{\alpha_0}{(1-\alpha_0)c_u} HH^T + \frac{\lambda_u}{(1-\alpha_0)c_u} I\Big)^{-1}\cdot H^{u}_S\cdot \mathbf{y}_S \\ 
        & \mathbf{h}_i^* \varpropto \Big(W^{i}_S{(W^{i}_S)}^T + \frac{\alpha_0}{(1-\alpha_0)c_u} 
        WW^T + \frac{\lambda_i}{(1-\alpha_0)c_u} I\Big)^{-1}\cdot W^{i}_S \cdot \mathbf{y}_S
    \end{split}
\end{equation*}
\end{scriptsize}
Interestingly by choosing the right $\alpha_0$ and $\lambda$, the above solution is in fact the same solution (up to constant factor) for the original $\mathcal{L}_{iALS}$ ~\cite{hu2008collaborative}. Following the above analysis and let 
$\alpha_0^\prime=\frac{\alpha_0}{(1-\alpha_0)c_u}$ and 
$\lambda^\prime=\frac{\lambda}{(1-\alpha_0)c_u}$
\end{proof}

\noindent{\bf Debiased EASE:}
EASE~\cite{ease} has shown to be a simple yet effective recommendation model and evaluation bases (for a small number of items), and it aims to minimize:
\begin{equation*}
    \begin{split}
        &\mathcal{L}_{ease}=||X-XW||^2_F + \lambda||W||^2_F\\
        & s.t. \quad diag(W)=0
    \end{split}
\end{equation*}
It has a closed-form solution~\cite{ease}:
\begin{equation}
\begin{split}
&P = (X^TX+\lambda I)^{-1}\\
    &W^* = I-P\cdot dMat(diag(1\oslash P))
\end{split}
\end{equation}
where $\oslash$ denotes the elementwise division, and $diag$ vectorize the diagonal and $dMat$ transforms a vector to a diagonal matrix. 

To apply the debiased MSE loss into the EASE objective, let us first further transform $\mathcal{L}^{Debiased}_{mse}$ into parts of known positive items and parts of unknown items: 
\begin{scriptsize}
\begin{equation*}
    \begin{split}
&\mathcal{L}^{Debiased}_{mse}=\sum\limits_{u} \Bigg[ \sum\limits_{i\in \mathcal{I}_u^+}[c_u(\hat{y}_{ui}-1)^2 -c_u\lambda \hat{y}^2_{uq}]+ \lambda\sum\limits_{t\in\mathcal{I}} \hat{y}_{ut}^2 \Bigg]\\
        &=\sum\limits_{u} \Bigg[ \sum\limits_{i\in \mathcal{I}_u^+}[c_u(\hat{y}_{ui}-1)^2 -c_u\lambda \hat{y}^2_{ui}]+ \lambda\sum\limits_{t\in\mathcal{I}_u^+} \hat{y}_{ut}^2 + \lambda\sum\limits_{p\in\mathcal{I}\backslash \mathcal{I}_u^+} \hat{y}_{ut}^2 \Bigg]\\
&=\sum\limits_{u} \Bigg[\sum\limits_{i\in \mathcal{I}_u^+}[c_u(\hat{y}_{ui}-1)^2+c_u \sum\limits_{p\in\mathcal{I}\backslash \mathcal{I}_u^+} \hat{y}_{ut}^2 \Bigg]\\
&+ \lambda (1-c_u)\sum\limits_{u}\sum\limits_{i\in \mathcal{I}_u^+}\hat{y}^2_{ui} + (\lambda-c_u) \sum\limits_{p\in\mathcal{I}\backslash \mathcal{I}_u^+} \hat{y}_{ut}^2 \\
 &=||\sqrt{C_u}(X-XW)||^2_F \\
        &-\lambda||X\odot \sqrt{C_u-I}XW||^2_F -||(1-X)\odot \sqrt{C_u-\lambda I}XW||^2_F
    \end{split}
\end{equation*}
\end{scriptsize}
Note that 
\begin{equation}
\sum\limits_{u}\sum\limits_{i\in \mathcal{I}}\hat{y}^2_{ui}=||XW||^2_F
\end{equation}

To find the closed-form solution, we further restrict $\lambda=1$ and consider $c_u$ as a constant (always $>1$), thus, we have the following simplified format: 


\begin{scriptsize}
\begin{equation}
    \begin{split}
    \mathcal{L}^{Debiased}_{mse}=&||\sqrt{C_u}(X-XW)||^2_F
        -||\sqrt{C_u-I}XW||^2_F\\
        &=||X-XW||^2_F
        -\alpha ||XW||^2_F 
    \end{split}
\end{equation}
\end{scriptsize}
where, $\alpha=c_u-1$. 
Note that if we only minimize this objective function $\mathcal{L}^{Debiased}_{mse}$, we have the closed form: 
$$W = ((1-\lambda )X^TX)^{-1}X^TX.$$ 

Now, considering the debiased version of EASE:  
\begin{equation*}
    \begin{split}
        &\mathcal{L}^{Debiased}_{ease}=||X-XW||^2_F-\alpha||XW||^2_F + \lambda||W||^2_F\\
        & s.t. \quad diag(W)=0
    \end{split}
\end{equation*}
It has the following closed-form solution (by similar inference as EASE~\cite{ease}):

\begin{equation}
\begin{split}
\widehat{W}&=\frac{1}{1-\alpha} (I -\hat{P}\cdot dMat( \vec{1}\oslash diag(\hat{P}) ), \\ 
where \ \  & \hat{P}= (X^TX+\frac{\lambda}{1-\alpha} I)^{-1}
\end{split}
\end{equation}
Now, we can make the following observation: 
\bthm
For any debiased EASE loss $\mathcal{L}^{Debiased}_{ease}$ with parameters $\alpha$ and $\lambda$ with constant $c_u>1$ for all users, there are original EASE loss with parameter,  $\lambda^\prime$, which have the same closed form solutions EASE (up to constant factor). 
\ethm 
\bproof
Following above analysis and let 
$\lambda^\prime=\frac{\lambda}{(1-\alpha)c_u}$.
\eproof 

These results also indicate the sampling based approach to optimize the debiased $MSE$ loss may also be rather limited. In the experimental results, we will further validate this conjecture on debiased MSE loss. 

\clearpage

\end{document}